\documentclass[a4paper,fleqn]{cas-dc}
\usepackage[switch]{lineno} 
\usepackage[numbers]{natbib}
\usepackage{graphicx}
\usepackage{subfigure}
\usepackage{amsmath}
\usepackage{amssymb}
\usepackage{amsthm}
\usepackage{booktabs}
\usepackage[ruled]{algorithm2e}
\usepackage{algpseudocode}
\usepackage{latexsym}
\usepackage{bm}

\usepackage{etoolbox}
\usepackage{pythonhighlight}

\def\tsc#1{\csdef{#1}{\textsc{\lowercase{#1}}\xspace}}
\tsc{WGM}
\tsc{QE}
\tsc{EP}
\tsc{PMS}
\tsc{BEC}
\tsc{DE}

\begin{document}

\let\WriteBookmarks\relax
\def\floatpagepagefraction{1}
\def\textpagefraction{.001}
\shortauthors{Zheng et~al.}

\title [mode = title]{Asymmetric double-winged multi-view clustering network for exploring Diverse and Consistent Information}

\author[1]{Qun Zheng}
\author[2]{Xihong Yang}
\author[2]{Siwei Wang}
\author[1]{Xinru An}
\author[1]{Qi Liu}[orcid=0000-0001-5454-2655]
\ead{qliu7@ustc.edu.cn}

\address[1]{School of Earth and Space Sciences, University of Science and Technology of China, Hefei 230001, China}
\address[2]{College of Computer Science and Technology, National University of Defense Technology, Changsha 410073, China}

\cortext[cor1]{Corresponding author at: School of Earth and Space Sciences, University of Science and Technology of China, Hefei 230001, China}

\begin{abstract}
In unsupervised scenarios, deep contrastive multi-view clustering (DCMVC) is becoming a hot research spot, which aims to mine the potential relationships between different views. Most existing DCMVC algorithms focus on exploring the consistency information for the deep semantic features, while ignoring the diverse information on shallow features. To fill this gap, we propose a novel multi-view clustering network termed CodingNet to explore the diverse and consistent information simultaneously in this paper. Specifically, instead of utilizing the conventional auto-encoder, we design an asymmetric structure network to extract shallow and deep features separately. Then, by aligning the similarity matrix on the shallow feature to the zero matrix, we ensure the diversity for the shallow features, thus offering a better description of multi-view data. Moreover, we propose a dual contrastive mechanism that maintains consistency for deep features at both view-feature and pseudo-label levels. Our framework's efficacy is validated through extensive experiments on six widely used benchmark datasets, outperforming most state-of-the-art multi-view clustering algorithms.
\end{abstract}





\maketitle

\section{Introduction}
Multi-view data is collected from different information sources, which is ubiquitous in many real-world applications, such as recommendation systems, multimedia analysis, and so on. Among those directions, multi-view clustering, aiming to divide the data into different clusters, has become a research hot spot. 

In recent years, lots of multi-view clustering methods \cite{chen1,chen2,chen3, chen4, DealMVC, suyuan_TNNLS} have been proposed, which can be roughly divided into five categories, i.e., non-negative matrix factorization methods (NMF) \cite{wenjie_mf}, graph-based clustering methods \cite{suyuan_AAAI,suyuan_TNNLS,siwei_nips}, multiple kernel clustering methods (MKC) \cite{one,simplemkkm,incomplete,efficient}, subspace clustering methods \cite{ZHOU_1,ZHOU_2,zhang2020adaptive}, and deep multi-view clustering methods \cite{MFLVC,SDSNE,COMIC,SDMVC}. To be specifically, non-negative matrix factorization methods utilize matrix factorization for the multi-view data. On the other hand, MKC effectively integrates information from multiple predefined kernel matrices to enhance clustering performance, facilitating better separation of complex data distributions. Graph-based clustering takes advantage of the multi-view data by constructing a unified graph structure. Meanwhile, subspace clustering methods focus on learning consistent subspace representations, allowing for the identification of shared subspaces across different views and promoting more accurate and robust clustering results. However, despite their individual strengths, these conventional multi-view clustering methods encounter challenges. The limited representation extraction capacity can restrict their ability to fully capture the intricacies of the data. Additionally, the high computation complexity of some of these algorithms may pose practical difficulties, leading to longer processing times and potential resource constraints.

\begin{figure}[t]
\centering
\includegraphics[scale=0.3]{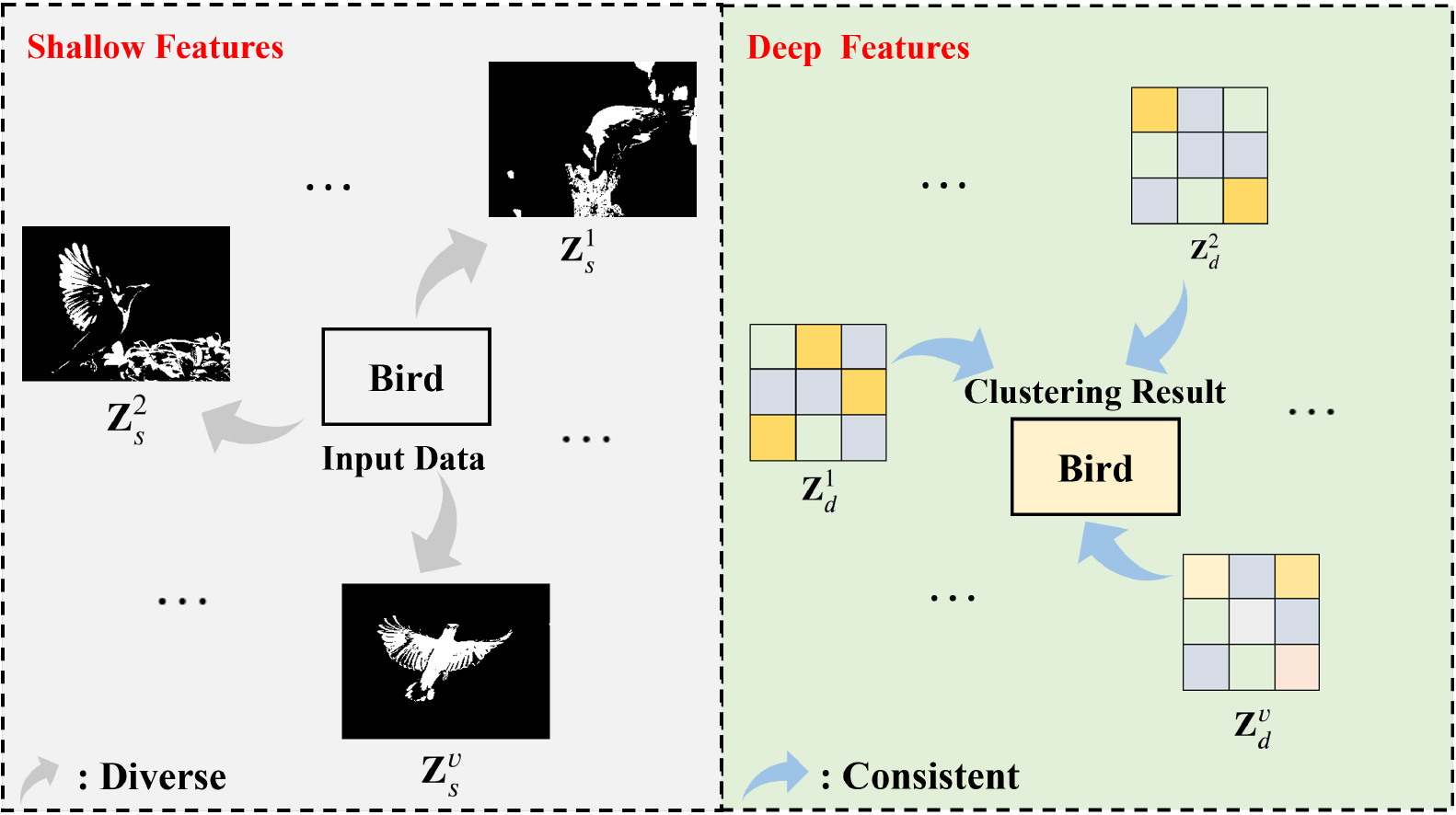}
\caption{An example illustration of the motivation. The shallow features $\textbf{Z}_s^v$ keep diversity, thereby capturing a wealth of information that enhances the portrayal of the features within the original multi-view data. In contrast, the deep features $\textbf{Z}_d^v$ are derived from intricate network mappings. These deep semantic representations assume the responsibility of retaining consistency, ensuring the preservation of category-related information across the multi-view data.}
\label{motivation}  
\end{figure}

Deep clustering methods have emerged as a promising solution to address the limited representation extraction capacity in traditional multi-view clustering \cite{dmvc1,dmvc3,dmvc4,Completer,EAMC,SiMVC}. Deep graph-based methods \cite{graph_m1,graph_m2} have been proposed to directly cluster multi-view data using affinity matrices. Moreover, adversarial multi-view methods \cite{adver_m1,adver_m2} employ generators and discriminators to align feature distributions for enhanced clustering performance. Furthermore, contrastive learning has recently gained popularity across various fields. COMPLETER \cite{Completer} utilizes contrastive learning to learn informative and consistent representations from multi-view data, while MFLVC \cite{MFLVC} adopts a contrastive strategy to learn different levels of features.

While numerous multi-view clustering techniques have showcased promising outcomes, they often center around fostering congruence in view features, inadvertently neglecting the critical facet of embracing diversity across varying perspectives. As depicted in Fig.\ref{motivation}, the shallow features should maintain diversity to faithfully capture the essence of the original multi-view data. In contrast, the deep features must exhibit consistency to yield consistent predictions across distinct view. For instance, the shallow features offer an initial characterization of the multi-view data. Their purpose is to amass a comprehensive gamut of information that mirrors the original dataset, thereby enhancing model training. In the case of the bird sample, the shallow features endeavor to capture diverse angles of the bird. Meanwhile, deep features entail distilled representations of the original data, housing profound semantic insights. Hence, preserving consistency is paramount to ensure consistent category predictions. Irrespective of the circumstances, the model must unfailingly recognize the input data as a bird rather than misclustering it as a cat.

In this paper, we design an asymmetri\textbf{C} d\textbf{o}uble-winge\textbf{d} multi-view cluster\textbf{ing} \textbf{net}work for clustering, named \textbf{CodingNet}. The overall framework of CodingNet is depicted in Fig.\ref{overall}. Initially, we devise an asymmetric network to capture shallow and deep view features separately. Subsequently, we introduce two modules to concurrently harness diverse and consistent information from multi-view data. Precisely, we enforce the similarity matrix of shallow features to align with a zero matrix, effectively reducing the inter-feature similarity at the shallow level. Simultaneously, in the latent space, we design a dual contrastive mechanism to ensure consistency among deep features across both view-feature and pseudo-label dimensions. Extensive experiments on six datasets have demonstrated the effectiveness and the superiority of our proposed algorithm.

The main contributions of this paper are listed as follows.

\begin{itemize}
\item To explore the diversity and consistency in multi-view clustering, we design an asymmetric double-winged network (CodingNet) to extract the shallow and deep view features, respectively.

\item By aligning the similarity matrix of shallow features with a zero matrix, we enhance the diversity of shallow features, thereby yielding a more comprehensive representation of the multi-view data. Moreover, we design a dual contrastive mechanism to learn the consistency of deep features on both view-feature and pseudo-label levels.

\item Our proposed algorithms exhibit robustness to hyperparameters. We substantiate the efficacy of our approach through ablation studies and visualization experiments. The extensive experimental results across six benchmark datasets underscore the superior performance and efficiency of our method.
\end{itemize}

\section{Related Work}

\subsection{Multi-view Clustering}

Multi-view Clustering (MVC) has gained significant attention in recent times. The existing MVC algorithms can be broadly categorized into two groups: traditional multi-view clustering methods and deep multi-view clustering methods. Traditional multi-view clustering methods encompass four primary categories. Firstly, there are matrix factorization-based algorithms \cite{mvc_mf, mvc_mf2, wenjie_mf}. These techniques employ non-negative matrix factorization to extract a common latent factor from the multi-view data. In the context of multi-view scenarios, \cite{mvc_mf3} proposed a shared clustering indicator matrix. Secondly, there is kernel learning-based MVC. In this approach, predefined kernels are employed to handle diverse views \cite{simplemkkm, one, incomplete}. By combining these kernels linearly or non-linearly, an integrated kernel is derived to facilitate clustering. Thirdly, graph-based MVC techniques \cite{CMC, graph_mvc,kang_graph,xia_graph,xia_graph1,xia_graph3} exploit the inherent structure within the multi-view data by constructing graphs that preserve relationships among the samples. Lastly, there are subspace-based MVC \cite{kang_subspace,xia_sub,suyuan_AAAI,suyuan_TNNLS} methods that focus on learning a consistent subspace representation for multi-view data. For instance, in \cite{subspace}, the authors introduced a diversity-induced mechanism for multi-view subspace clustering. Nevertheless, the aforementioned conventional approaches could merely mine the shallow representation of the multi-view data. As a result, the representation lacks the necessary discriminative capacity.

Deep neural networks \cite{zhang1,zhang2,zhang3,zhang4,MGCN} have demonstrated their capability in extracting robust feature representations, which has led to the development of deep multi-view clustering methods in recent years \cite{dmvc1, dmvc4, dmvc3}. These methods can be categorized into two groups: one-stage algorithms \cite{DAMC, EAMC} and two-stage algorithms \cite{MultiVAE, Completer}. Leveraging the power of deep learning, deep multi-view clustering algorithms have the ability to uncover latent cluster patterns within multi-view data.

\subsection{Contrastive Learning}

Contrastive learning has garnered significant attention in both computer vision \cite{DIM, SIMCLR, BYOL, BARLOW, GCC, xihong} and graph clustering \cite{DCRN, SCGC, CCGC, HSAN, GCC-LDA, CONVERT} due to its robust ability to extract inherent supervisory information. The core tenet of contrastive learning revolves around amplifying the similarity between positive samples while minimizing it between negative samples in the latent space. Noise contrastive estimation (NCE) \cite{nce1, nce2} initially proposed this principle. InfoNCE \cite{infonce} subsequently expanded NCE to differentiate between distinct views of a sample. Building on these concepts, Moco \cite{MOCO} and SimCLR \cite{SIMCLR} pioneered image-wise feature learning by drawing positive sample pairs closer and pushing negative ones apart.

In the realm of multi-view clustering, various contrastive learning techniques have been proposed. For instance, CMC \cite{CMC} introduced a multi-view coding framework for contrastive learning, aimed at extracting underlying semantic insights. MVGRL \cite{MVGRL} leveraged graph diffusion matrices to create augmented graphs, implementing a multi-view contrastive approach for downstream tasks. More recently, MFLVC \cite{MFLVC} introduced dual objectives for multi-view clustering via contrastive learning, emphasizing both high-level features and semantic labels.

\begin{figure*}
\centering
\includegraphics[scale=0.65]{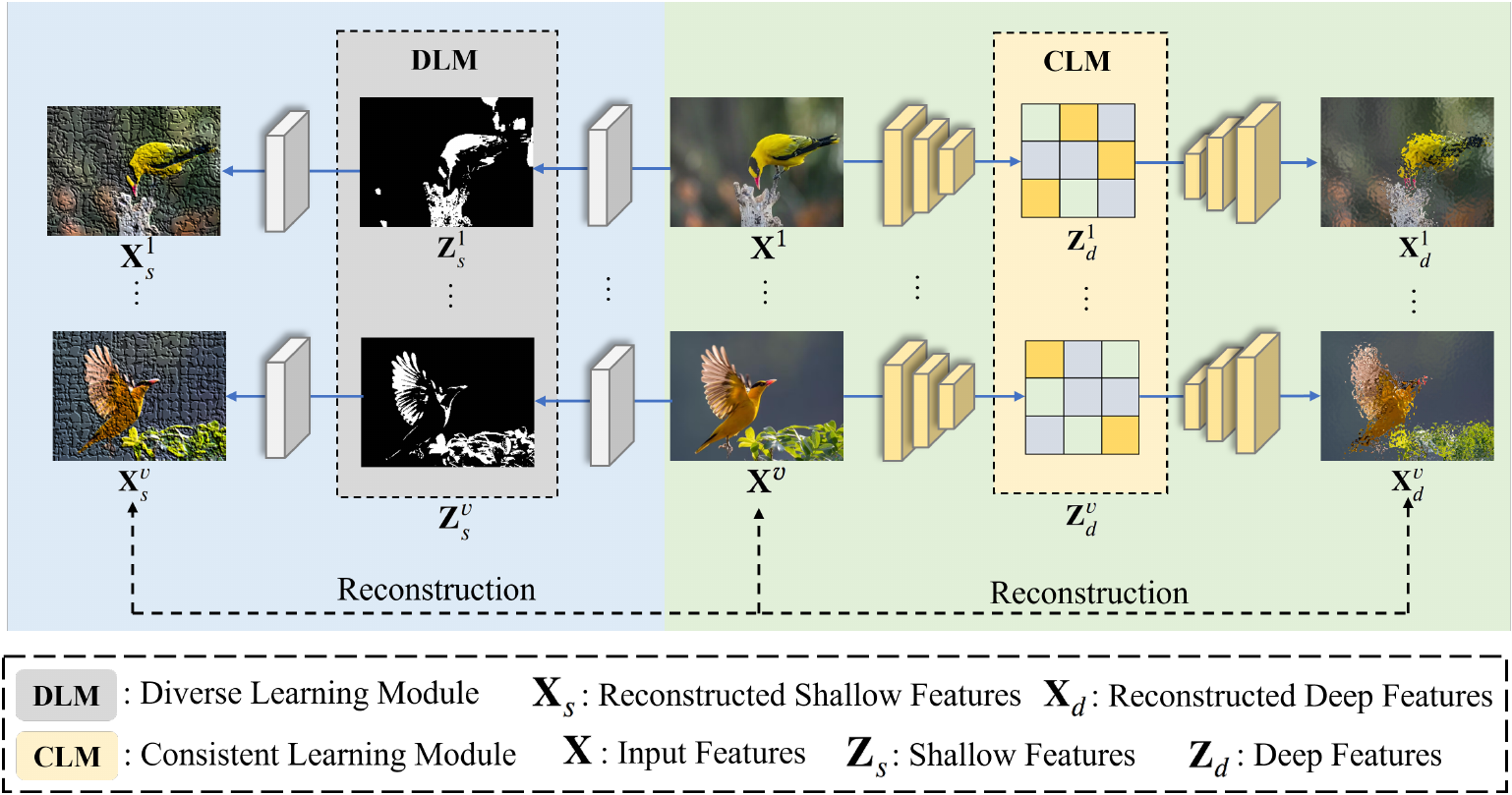}
\caption{Illustration of our proposed asymmetric double-winged multi-view clustering network. In our approach, we extract shallow and deep view features using two distinct encoder network structures. Subsequently, we introduce two modules: the Diverse Learning Module ($\textbf{DML}$) and the Consistent Learning Module ($\textbf{CLM}$). These modules enable us to explore diverse and consistent information, respectively. Specifically, we align the similarity matrix of shallow features with a zero matrix, effectively enhancing the diversity of shallow features. Additionally, we introduce a dual contrastive mechanism that ensures the consistency of deep features at both the view-feature and pseudo-label levels.}
\label{overall}   
\end{figure*}

\section{Method}

In this section, we propose a novel multi-view clustering algorithm, termed CodingNet. The overall framework of CodingNet is shown in Fig. \ref{overall}. 

\subsection{Notations and Preliminary}

We denote the multi-view data as $\textbf{X} = {\textbf{X}^1, \textbf{X}^2, \dots, \textbf{X}^v} \in \mathbb{R}^{N \times D_v}$, where $\textbf{X}^v$ represents the samples in the $v$-th view, and $D_v$ represents the dimensionality of the samples in that view. Furthermore, $N$ denotes the number of samples in each view. The objective of multi-view clustering is to partition the samples into $K$ clusters.

Inspired by the strong ability of the feature extraction, autoencoder has attracted the attention of researchers \cite{autoencoder1, autoencoder2}. In our approach, we define $f(\theta^v;\cdot)$ and $g(\xi^v; \cdot)$ as encoder network and decoder network, respectively. Therefore, we could obtain the view feature and the reconstruction feature as:

\begin{equation}
\begin{aligned}
\textbf{Z}^v &= f(\theta^v; \textbf{X}^v),\\
\tilde{\textbf{X}}^v &= g(\xi^v; \textbf{Z}^v) = g(\xi^v;(f(\theta^v;\textbf{X}^v)))
\end{aligned}
\end{equation}
where $\textbf{Z}^v$ is the view feature, ${\tilde{\textbf{X}}^v}$ is the reconstruction feature. Through the autoencoder, we optimize the reconstruction loss to learn the latent view feature by minimizing:

\begin{equation} 
\mathcal{L}_{rec} = \sum_{v=1}^V  \left \| \textbf{X}^v - g(\xi^v;(f(\theta^v;\textbf{X}^v))) \right \|_F^2.
\label{auto_loss}
\end{equation}

\subsection{Diverse Learning Module} 

\begin{figure}[t]
\centering
\includegraphics[scale=0.3]{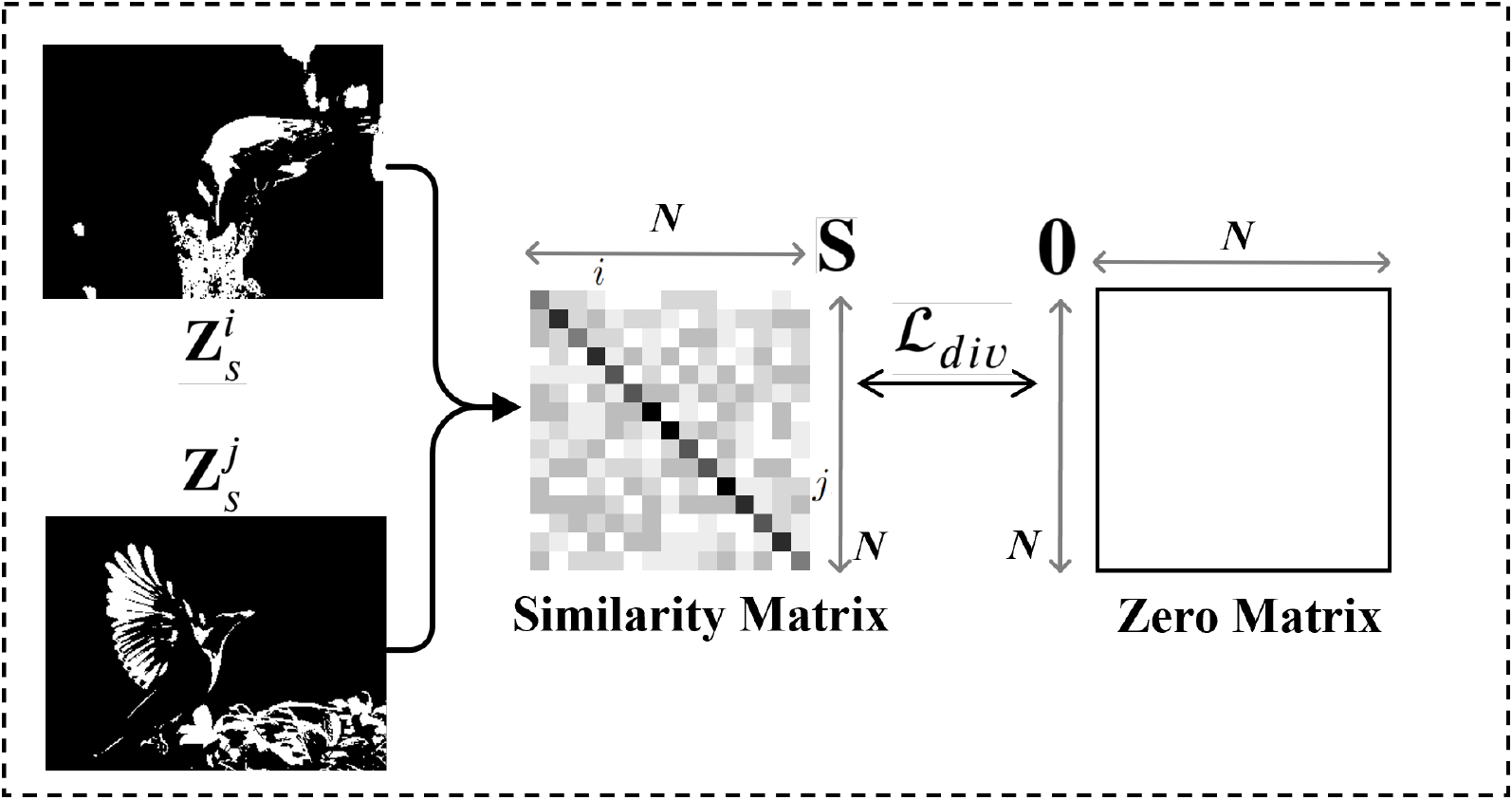}
\caption{Illustration of diverse learning module (DLM). By aligning the similarity matrix to the zero matrix, we reduce the similarity for the shallow view features, thus keeping the diversity of the multi-view data.}
\label{DLM}  
\end{figure}

Diverse information is a piece of important information in multi-view data. In this subsection, we design a diverse learning module (DLM) to explore the shallow feature in multi-view scenarios. The framework of DLM is shown in Fig.\ref{DLM}. Specifically, a simple encoder network is proposed to obtain the shallow view features as follows:

\begin{equation}
\begin{aligned}
\textbf{Z}^v_s = f_s(\theta^v; \textbf{X}^v),
\end{aligned}
\label{simple_encoder}
\end{equation}
where $f_s(\theta^v;\cdot)$ signifies the simple encoder network. For this purpose, we employ a single-layer Multi-Layer Perceptron (MLP) as $f_s(\theta^v;\cdot)$ to capture initial view features. Subsequently, we utilize $\ell_2$ to normalize $f_s(\theta^v;\cdot)$:

\begin{equation}
\begin{aligned}
\textbf{Z}^v_s = \frac{\textbf{Z}^v_s}{||\textbf{Z}^v_s||_2}.
\end{aligned}
\end{equation}

By this manner, we could obtain the shallow view features $\textbf{Z}^v_s$, which are extracted by the simple single-layer MLP. In order to explore the diversity of multi-view data, the shallow features should keep diversity across various views, reflecting the inherent traits of the input data. The process of preserving diversity in the shallow view features can be divided into two steps.

Firstly, the similarity matrix between two shallow features is calculated as:
\begin{equation}
\textbf{S}_{ij} =\frac{(\textbf{Z}_{i} )(\textbf{Z}_{j} )^{T} }{||\textbf{Z}_{i} ||_2||\textbf{Z}_{j} ||_2}.
\label{sim_s}
\end{equation}

Here, $\textbf{S}_{ij}$ is calculated by cosine similarity. After that, we force the similarity matrix to be equal to a zero matrix by minimizing the following formula:

\begin{equation}
\begin{aligned}
\mathcal{L}_{div} &= \frac{1}{N^2}\sum (\textbf{S}-\textbf{0})^2.
\end{aligned}
\label{correlation_loss}
\end{equation}

To elaborate, the diagonal elements signify the representation of the same sample in different views. By aligning the similarity of these diagonal elements to 0, we ensure the retention of diversity among the shallow features. Conversely, the off-diagonal elements represent different samples across the views, making it logical to minimize their similarity towards 0.

\subsection{Consistent Learning Module}

\begin{figure}[t]
\centering
\includegraphics[scale=0.3]{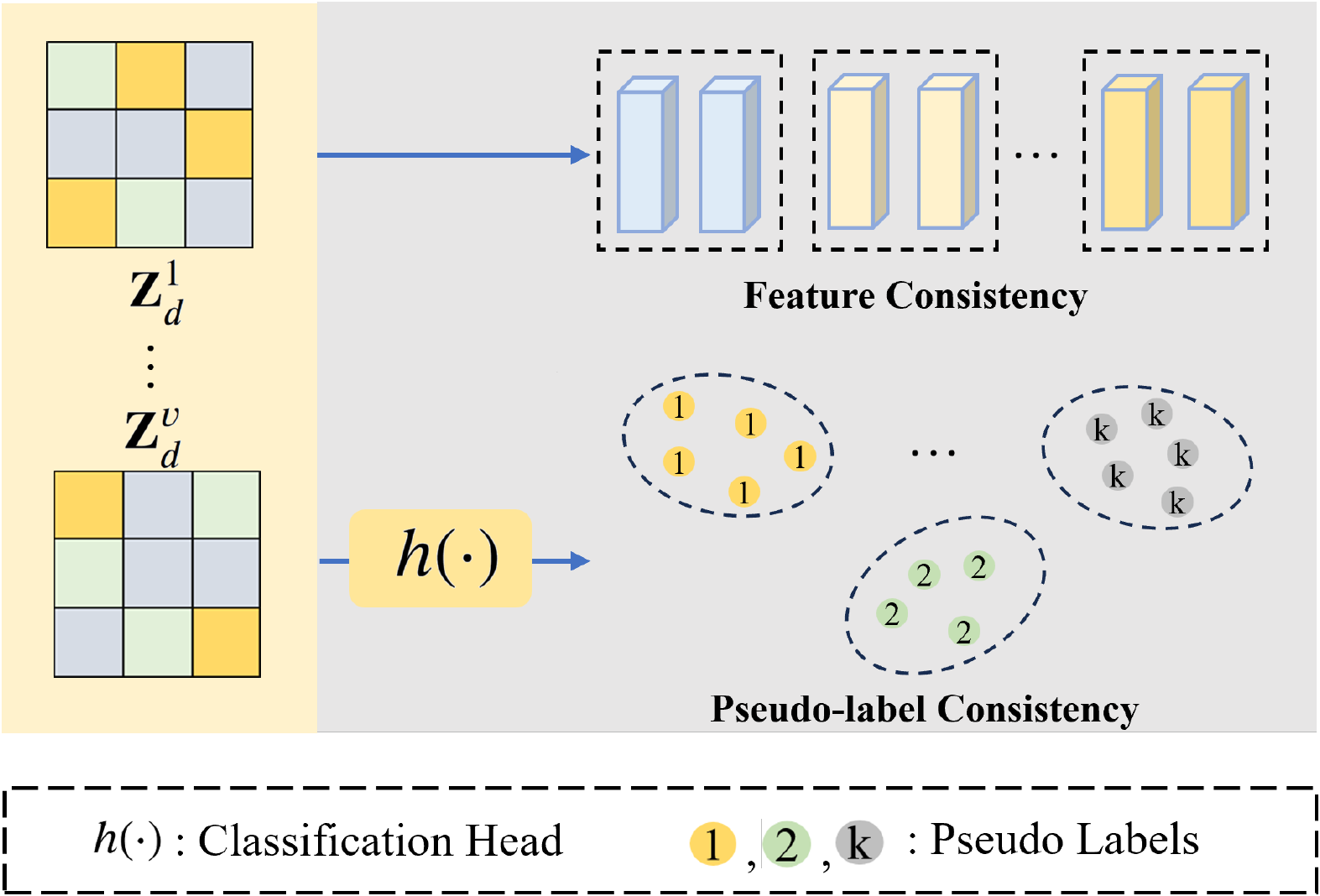}
\caption{Illustration of the consistent learning module (CLM). To be concrete, we design a dual contrastive mechanism to learn the consistency in both view-feature level and pseudo-label levels.}
\label{CLM}  
\end{figure}

Contrastive learning often focuses on extracting consistent information from multi-view data. Illustrated in Fig. \ref{CLM}, our consistent learning module (CLM) is designed to facilitate this. We employ a sophisticated encoder network, denoted as $f_d(\cdot;\theta^v)$, to derive the deep view feature $\textbf{Z}^v_d$:

\begin{equation}
\begin{aligned}
\textbf{Z}^v_d &= f_d(\theta^v; \textbf{X}^v),\\
\textbf{Z}^v_d &= \frac{\textbf{Z}^v_d}{||\textbf{Z}^v_d||_2}.
\end{aligned}
\label{complex_encoder}
\end{equation}

Then, we could obtain the pseudo labels by the classification head $h(\cdot)$ in an end-to-end manner. Specifically, the deep view features are fed into an MLP layer. The cluster alignments for all views are generated by:

Subsequently, the pseudo labels can be derived through the classification head $h(\cdot)$ in an end-to-end manner, generating cluster alignments for all views:

\begin{equation}
\begin{aligned}
\textbf{P}^v_d &= h(\textbf{Z}^v_d),
\label{pseudo_labels}
\end{aligned}
\end{equation}
where $\textbf{P}^v_{d}$ is the output probability of the deep view features.

The contrastive loss for the for deep view feature is formulated as:
\begin{equation}
\begin{aligned}
\ell(i) = -log \frac{\text{e}^{(\text{sim}(\textbf{Z}_i^m, \textbf{Z}_i^n)/\tau)}}{\sum_{k=1}^{N}{ \mathbb I_{\left \lfloor k \neq i \right \rfloor } \text{e}^{((\text{sim}(\textbf{Z}_i^m, \textbf{Z}_k^n)/\tau)}}}\\
-log \frac{\text{e}^{((\text{sim}(h(\textbf{Z}_i^m), h(\textbf{Z}_i^n))/\tau)}}{\sum_{k=1}^{N}{ \mathbb I_{\left \lfloor k \neq i \right \rfloor } \text{e}^{((\text{sim}(h(\textbf{Z}_i^m), h(\textbf{Z}_k^n)/\tau))}}},
\end{aligned}
\end{equation}
where $\mathbb I_{\lfloor {k\neq i} \rfloor} \in {0,1}$ is an indicator function. $m,n \in \{1, \dots, N\}$. $\tau$ is the temperature parameter. $(\text{sim}(\cdot)$ denotes the similarity between two deep view features. The deep feature contrastive loss across all views as:

\begin{equation}
\begin{aligned}
\mathcal{L}_{con}=\frac{1}{N} \sum_{i=1}^{N}\ell(i) 
\label{con_loss}
\end{aligned}
\end{equation}

In our paper, we ensure diversity by compelling the similarity matrix of shallow view features to approach zero matrices. This approach leads to a decrease in the similarity between any two shallow view features during training. Simultaneously, we utilize the contrastive loss to maintain consistency among deep view features, achieving this by minimizing the distance between positive samples and increasing the distance from negative samples.

\subsection{Loss function}

Our proposed CodingNet simultaneously optimizes four objectives, including the reconstruction loss $\mathcal{L}_{rec}$, the diversity loss $\mathcal{L}_{div}$, the consistent loss $\mathcal{L}_{con}$, and the consistent prediction loss $\mathcal{L}_{cp}$.

In detail, $\mathcal{L}_{cp}$ is designed to keep the same prediction for different views:

\begin{equation}
\begin{aligned}
\mathcal{L}_{cp} &=\sum_{v=1}^V{||\textbf{P}^v_{d} -\textbf{P}^v_{s} ||^{2} _{F}}
\label{cp_loss}
\end{aligned}
\end{equation}
where $||\cdot||$ is the mean squared error loss. $\textbf{P}_{d}$ and $\textbf{P}_{s}$ are the prediction results of the deep and shallow view features, respectively. Overall, the loss function of CodingNet is formulated as:
\begin{equation}
\mathcal{L} = \mathcal{L}_{rec}+ \mathcal{L}_{div} + \mathcal{L}_{con} + \mathcal{L}_{cp}
\label{LOSS}
\end{equation}
The detailed learning procedure of CodingNet is illustrated in Algorithm \ref{A1}.

\begin{algorithm}[!t]
  \caption{CodingNet}
  \label{A1}
  \LinesNumbered
  \KwIn{Raw features: $\{\mathbf{X}^{(v)}\}_{v=1}^V$, the interaction number $E$}
  \KwOut{Clustering result $\textbf{R}$.}
  Initialize the designed modules, including diverse learning module, and consistent learning module\;
  \While{$e=1$ < $E$}
  {
  Obtain the shallow view features $\textbf{Z}^v_s$ and $\textbf{Z}^v_d$ by the simple encoder network and complex encoder network with Eq.\eqref{simple_encoder} and Eq.\eqref{complex_encoder}, respectively;

  Calculate the similarity matrix ${\textbf{S}}$ for shallow view features with Eq.\eqref{sim_s};

  Force ${\textbf{S}}$ to approximate an identity matrix and calculate diverse feature learning loss \textit{$\mathcal{L}_{div}$} with Eq. \eqref{correlation_loss};

  Generate the pseudo labels by Eq.\eqref{pseudo_labels};

  Calculate the consistent feature learning loss $\mathcal{L}_{con}$ with Eq.\eqref{con_loss};

  Calculate the consistent prediction loss $\mathcal{L}_{cp}$ with Eq. \eqref{cp_loss};

  Update the whole network by minimizing $\mathcal{L}$ in Eq. \eqref{LOSS};      
  }

  \textbf{return:} Output the clustering result ${\textbf{R}}$.
\end{algorithm}

\section{Experiments}

\begin{table}[]
\centering
\caption{\textbf{Statistics summary of six datasets.}}
\scalebox{1.15}{
\begin{tabular}{cccc}
\hline
Dataset      & Samples & Clusters & Views \\ \hline
Reuters      & 1200    & 6        & 5     \\
CALTECH101\_7 & 1400    & 7        & 5     \\
UCI\_digit    & 2000    & 10       & 3     \\
CALTECH101   & 9144    & 102      & 5     \\
Hdigit       & 10000   & 10       & 2     \\
STL10\_4Views & 13000   & 10       & 3     \\ \hline
\end{tabular}}
\label{DATASET_INFO}
\end{table}

In this section, we conduct experiments to verify the effectiveness of our proposed CodingNet. The experiments aim to address the following key questions: 

\begin{itemize}
\item \textbf{RQ1}: How does CodingNet perform in multi-view clustering?
\item \textbf{RQ2}: What influence does the proposed module have on the performance of CodingNet?
\item \textbf{RQ3}: How do the hyper-parameters affect the performance of CodingNet?
\item \textbf{RQ4}: What clustering structure does CodingNet reveal?
\end{itemize}

\subsection{Experimental Setup}

We conducted the experiments on a desktop computer equipped with an Intel Core i7-7820x CPU, an NVIDIA GeForce RTX 2080Ti GPU, 64GB of RAM, and utilized the PyTorch deep learning platform. In all experiments, we trained the methods until convergence, using the Adam optimizer \cite{ADAM} to minimize the total loss.

\textbf{Datasets:} The datasets used in our paper include Reuters\footnote{\url{http://archive.ics.uci.edu/ml/datasets/Reuters-21578+Text+Categorization+Collection}}, Caltech101\_7\footnote{\url{https://data.caltech.edu/records/mzrjq-6wc02}}, UCI\_dighit\footnote{\url{https://cs.nyu.edu/∼roweis/data.html}}, Hdigit\footnote{\url{https://cs.nyu.edu/∼roweis/data.html}}, Caltech101\footnote{\url{https://data.caltech.edu/records/mzrjq-6wc02}}, and STL10\footnote{\url{https://cs.stanford.edu/~acoates/stl10/}}. The brief information of these datasets is summarized in Table \ref{DATASET_INFO}.

\textbf{Evaluation Metrics:} In this paper, we adopt the widely used metrics in multi-view clustering, i.e., accuracy (ACC), normalized mutual information (NMI), and purity (PUR)\cite{ZHOU_1,siwei_1,siwei_2}.

\textbf{Parameter Setting:}
During training, the learning rate is set to 0.0003, and a batch size of 256 is used.

\begin{table*}[]
\centering
\caption{Performance comparison across eight benchmark datasets in the context of multi-view clustering (Part 1/2). The optimal outcomes are indicated in \textbf{bold}, while the second-best values are \underline{underlined}.}
\scalebox{1.15}{
\begin{tabular}{cc|ccc|ccc|ccc}
\hline
\multicolumn{2}{c|}{Methods}                 & \multicolumn{3}{c|}{Reuters}                     & \multicolumn{3}{c|}{Caltech101\_7}               & \multicolumn{3}{c}{UCI\_digit}                   \\ \hline
\multicolumn{2}{c|}{Evaluation Metrics (\%)} & ACC            & NMI            & PUR            & ACC            & NMI            & PUR            & ACC            & NMI            & PUR            \\ \hline
AE2-Nets             & CVPR 2019             & 22.55          & 03.48          & 24.42          & 49.72          & 36.09          & 52.05          & 80.57          & 69.51          & 80.57          \\
COMIC                & ICML 2019             & 16.67          & 12.57          & 16.67          & 53.20          & 54.90          & 60.40          & 73.55          & 81.80          & 81.35          \\
DEMVC                & IS 2020               & \underline{45.00}    &  \underline{21.08}    &  \underline{45.08}    & 54.86          & 41.77          & 56.00          & 62.30          & 60.54          & 63.15          \\
SDMVC                & TKDE 2021             & 17.67          & 14.73          & 18.58          & 44.21          & 30.97          & 47.93          & 63.00          & 64.29          & 67.05          \\
CoMVC                & CVPR 2021             & 32.25          & 13.58          & 33.83          & 40.84          & 28.66          & 69.95          & 46.4           & 49.36          & 46.45          \\
SiMVC                & CVPR 2021             & 33.58          & 10.36          & 33.67          & 75.03          & 37.07          & 77.54          & 20.75          & 16.59          & 21.75          \\
SDSNE                & AAAI 2022             & 23.25          & 20.28          & 27.00          & 76.14          & \textbf{74.52} & 78.93          & 84.55          & \textbf{89.05} & 87.05          \\
MFLVC                & CVPR 2022             & 39.92          & 20.01          & 41.42          &  \underline{80.40}    & 70.30          &  \underline{80.40}    &  \underline{92.00}    & 85.40          &  \underline{92.00}    \\
DSMVC                & CVPR 2022             & 43.83          & 18.11          & 45.00          & 62.64          & 47.34          & 64.43          & 85.45          & 80.67          & 85.40          \\ \hline
CodingNet                  & Ours                  & \textbf{45.83} & \textbf{26.02} & \textbf{46.00} & \textbf{83.71} &  \underline{73.47}    & \textbf{83.71} & \textbf{93.25} &  \underline{87.74}    & \textbf{93.25} \\ \hline
\end{tabular}}
\label{com_exp1}
\end{table*}

\begin{table*}
\centering
\caption{Performance comparison across eight benchmark datasets in the context of multi-view clustering (Part 1/2). The optimal outcomes are indicated in \textbf{bold}, while the second-best values are \underline{underlined}. The abbreviation O/M stands for out-of-memory error experienced during training.}
\scalebox{1.15}{
\begin{tabular}{cc|ccc|ccc|ccc}
\hline
\multicolumn{2}{c|}{Methods}            & \multicolumn{3}{c|}{CALTECH101}                 & \multicolumn{3}{c|}{Hdigit}                    & \multicolumn{3}{c}{STL10}                        \\ \hline
\multicolumn{2}{c|}{Evaluation Metrics (\%)} & ACC           & NMI            & PUR            & ACC           & NMI            & PUR           & ACC            & NMI            & PUR            \\ \hline
AE2-Nets           & CVPR 2019          & 13.95         & \underline{30.54}    & 25.43          & 83.74         & 86.56          & 90.81         & 9.60            & 17.76          & 23.19          \\
COMIC              & ICML 2019          & 17.22         & 26.88          & 20.67          & 94.21         & 86.96          & 94.21         & 18.27          & 11.76          & 18.71          \\
DEMVC              & IS 2020            & 11.05         & 22.84          & 19.66          & 38.65         & 37.08          & 42.17         & 28.34          & \underline{26.19}     & 30.13          \\
SDMVC              & TKDE 2021          & 15.11         & 30.48          & \underline{28.78}    & 27.86         & 23.09          & 28.13         & 30.01          & 25.27          & 31.07          \\
CoMVC              & CVPR 2021          & 16.36         & 25.61          & 23.71          & 34.63         & 39.81          & 36.35         & 23.55          & 16.26          & 25.01          \\
SiMVC              & CVPR 2021          & 13.48         & 18.18          & 17.91          & 23.48         & 11.20           & 25.02         & 16.04          & 6.34           & 17.02          \\
SDSNE              & AAAI 2022          & O/M           & O/M            & O/M            & O/M           & O/M            & O/M           & O/M            & O/M            & O/M            \\
MFLVC              & CVPR 2022          & \textbf{21.30} & 28.60           & 28.23          & \underline {98.82}   & \underline {98.41}    & \underline{ 98.82}   & \underline{31.14}    & 25.36          & \underline{31.25}    \\
DSMVC              & CVPR 2022          & 16.27         & 26.53          & 25.60           & 98.06         & 97.36          & 98.06         & 27.53          & 19.33          & 29.41          \\ \hline
CodingNet                & Ours               & \underline{19.05}   & \textbf{36.79} & \textbf{34.28} & \textbf{99.7} & \textbf{99.06} & \textbf{99.7} & \textbf{35.77} & \textbf{29.02} & \textbf{35.97} \\ \hline
\end{tabular}}
\label{com_exp2}
\end{table*}

\begin{table*}[]
\centering
\caption{Ablation Studies of CodingNet in six datasets. ``(w/o) D'', ``(w/o) C'', ``(w/o) P'', and  ``(w/o) D\&C\&P'' represent the reduced models by removing the diverse learning module, the consistent learning module, the consistent prediction module, and all designed modules, respectively. Besides, ``Sym'' denotes the symmetric network structure. ``B-D'' and ``B-C'' mean that the network is optimized by separately exploring diversity and consistency.}
\begin{tabular}{c|c|cccc|ccc|c}
\hline
\textbf{Dataset}                        & \textbf{Metrics} & \textbf{(w/o)D} & \textbf{(w/o)C} & \textbf{(w/o)P} & \textbf{(w/o)D\&C\&P} & \textbf{Sym} & \textbf{B-D} & \textbf{B-C} & \textbf{Ours} \\ \hline
\multirow{3}{*}{\textbf{Reuters}}       & \textbf{ACC}     & 41.75           & 26.25           & 40.25           & 24.50                 & 36.75        & 16.67          & 41.67          & 45.83         \\
                                        & \textbf{NMI}     & 24.59           & 05.06           & 25.31           & 08.76                 & 22.24        & 00.00          & 25.03          & 26.02         \\
                                        & \textbf{PUR}     & 41.92           & 26.25           & 43.08           & 24.83                 & 40.25        & 16.67          & 41.83          & 46.00         \\ \hline
\multirow{3}{*}{\textbf{Caltech101\_7}} & \textbf{ACC}     & 77.50           & 26.50           & 81.71           & 23.14                 & 75.00        & 14.29          & 76.43          & 83.71         \\
                                        & \textbf{NMI}     & 70.42           & 07.60           & 72.36           & 09.95                 & 68.20        & 00.00          & 68.00          & 73.47         \\
                                        & \textbf{PUR}     & 70.42           & 26.57           & 81.71           & 23.14                 & 75.00        & 14.29          & 76.43          & 83.71         \\ \hline
\multirow{3}{*}{\textbf{UCI\_digit}}    & \textbf{ACC}     & 91.95           & 20.95           & 84.30           & 12.95                 & 83.85        & 10.00          & 82.95          & 93.25         \\
                                        & \textbf{NMI}     & 85.83           & 10.80           & 82.14           & 05.22                 & 80.48        & 00.00          & 80.27          & 87.74         \\
                                        & \textbf{PUR}     & 91.95           & 20.95           & 84.30           & 13.50                 & 83.85        & 10.00          & 82.95          & 93.25         \\ \hline
\multirow{3}{*}{\textbf{CALTECH101}}    & \textbf{ACC}     & 15.87           & 10.33           & 14.86           & 09.56                 & 14.74        & 08.75          & 16.19          & 19.05         \\
                                        & \textbf{NMI}     & 36.01           & 01.84           & 35.99           & 09.56                 & 35.70        & 00.00          & 36.21          & 36.79         \\
                                        & \textbf{PUR}     & 33.56           & 10.35           & 33.71           & 09.69                 & 33.50        & 08.75          & 34.04          & 34.28         \\ \hline
\multirow{3}{*}{\textbf{Hdigit}}        & \textbf{ACC}     & 89.28           & 12.51           & 98.41           & 19.31                 & 14.59        & 14.93          & 98.39          & 99.70         \\
                                        & \textbf{NMI}     & 89.20           & 02.24           & 95.65           & 07.94                 & 04.71        & 10.09          & 95.67          & 99.06         \\
                                        & \textbf{PUR}     & 89.28           & 12.51           & 98.41           & 19.68                 & 14.59        & 14.93          & 98.39          & 99.70         \\ \hline
\multirow{3}{*}{\textbf{STL10}}         & \textbf{ACC}     & 32.67           & 13.27           & 17.56           & 11.75                 & 32.62        & 10.00          & 33.98          & 35.77         \\
                                        & \textbf{NMI}     & 27.24           & 01.72           & 08.12           & 01.02                 & 26.87        & 00.00          & 27.94          & 29.02         \\
                                        & \textbf{PUR}     & 33.92           & 13.27           & 17.88           & 11.77                 & 33.72        & 10.00          & 35.34          & 35.97         \\ \hline
\end{tabular}
\label{ablation_studies}
\end{table*}

\begin{figure*}[]
\centering
\small
\begin{minipage}{0.27\linewidth}
\centerline{\includegraphics[width=1\textwidth]{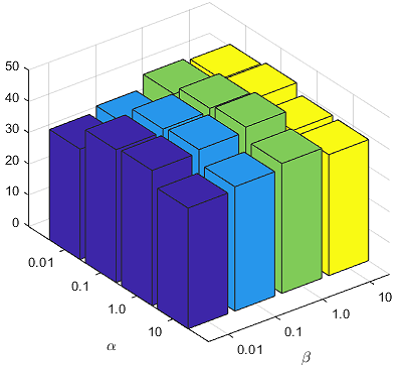}}\hspace{10mm}
\vspace{3pt}
\textbf{\centerline{Reuter-ACC}}
\centerline{\includegraphics[width=\textwidth]{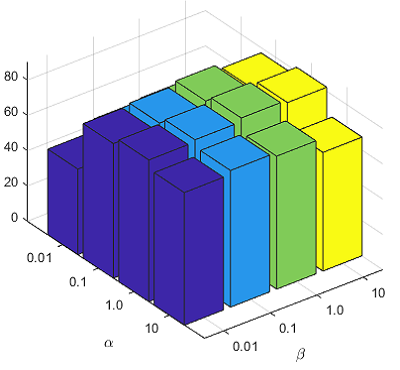}}\hspace{10mm}
\vspace{3pt}
\textbf{\centerline{Caltech101\_7-ACC}}
\end{minipage}
\begin{minipage}{0.27\linewidth}
\centerline{\includegraphics[width=\textwidth]{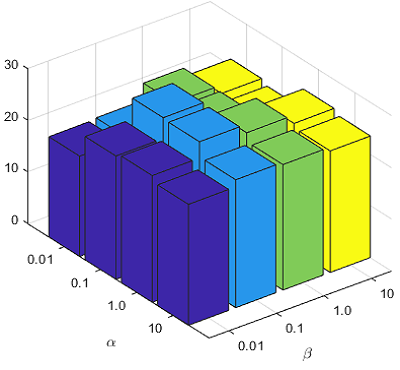}}\hspace{10mm}
\vspace{3pt}
\textbf{\centerline{Reuter-NMI}}
\centerline{\includegraphics[width=1\textwidth]{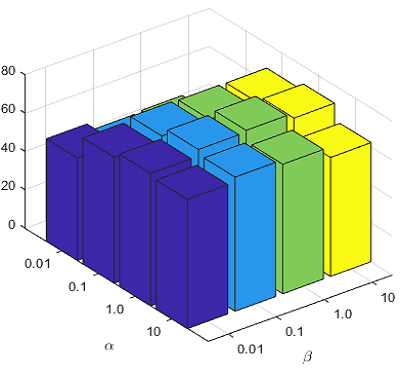}}\hspace{10mm}
\vspace{3pt}
\textbf{\centerline{Caltech101\_7-NMI}}
\end{minipage}
\begin{minipage}{0.27\linewidth}
\centerline{\includegraphics[width=\textwidth]{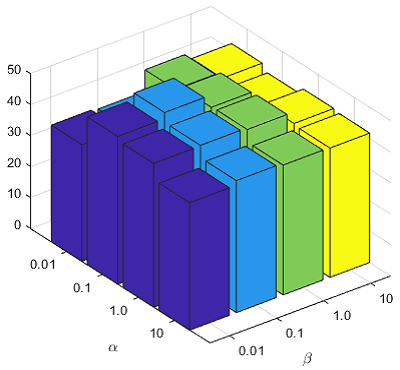}}\hspace{10mm}
\vspace{3pt}
\textbf{\centerline{Reuter-PUR}}
\centerline{\includegraphics[width=\textwidth]{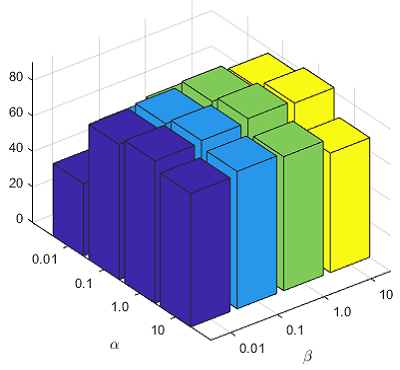}}\hspace{10mm}
\vspace{3pt}
\textbf{\centerline{Caltech101\_7-PUR}}
\end{minipage}
\caption{Sensitivity analysis of the hyper-parameter $\alpha$ and $\beta$ on two datasets with three metrics.}
\label{Sensitivity}
\end{figure*}

\textbf{Comparison methods}

The proposed CodingNet algorithm is compared against a total of seven deep multi-view clustering algorithms. These clustering algorithms can be roughly categorized into two groups: deep multi-view clustering algorithms (AE2-Net \cite{AE2Net}, CoMIC \cite{COMIC}, DEMVC \cite{DEMVC}, SDMVC \cite{SDMVC}, SDSNE \cite{SDSNE}, DSMVC \cite{DSMVC}, SiMVC \cite{SiMVC}) and contrastive multi-view clustering algorithms (CoMVC \cite{SiMVC}, MFLVC \cite{MFLVC}).

\subsection{Performance Comparison~\textbf{(RQ1)}}

In this subsection, we implement experiments to demonstrate the effectiveness of the proposed CodingNet. To be concrete, two categories deep multi-view algorithms are compared our method with nine baselines, including classical deep multi-view clustering algorithms ((AE2-Nets \cite{AE2Net}, COMIC \cite{COMIC}, DEMVC \cite{DEMVC}, SDMVC \cite{SDMVC}, SDSNE \cite{SDSNE}, DSMVC \cite{DSMVC}, CoMVC \cite{SiMVC})), and deep contrastive multi-view clustering algorithms (SiMVC \cite{SiMVC}, MFLVC \cite{MFLVC}).

The results are shown in Table.\ref{com_exp1} and Table. \ref{com_exp2}. We could obtain the following observations.
\begin{enumerate}
  \item Our proposed CodingNet could achieve promising clustering performance compared with the classical deep multi-view clustering algorithms, e.g., COMIC \cite{COMIC}, DEMVC \cite{DEMVC}.  This improvement can be attributed to the incorporation of contrastive learning in CodingNet, which helps in generating more discriminative features. Taking the result on the Reuters dataset for example, our CodingNet exceeds the runner-up method by 0.83\%, 4.94\%, and 0.92\% increments with respect to ACC, NMI, and PUR.

  \item When compared to deep contrastive multi-view clustering algorithms like SiMVC \cite{SiMVC} and MFLVC \cite{MFLVC}, CodingNet consistently exhibits superior performance. We speculate that the reason may be that the traditional contrastive algorithms focus on learning the consistency of the view feature, while ignoring the diversity.

  \item Our proposed method exhibits excellent scalability, making it suitable for handling large-scale datasets such as STL\_10. From the results in Table.\ref{com_exp2}, our CodingNet surpasses the runner-up by CodingNet in terms of ACC, CodingNet in terms of NMI, and CodingNet in terms of PUR.

\end{enumerate}

Overall, the aforementioned observations unequivocally establish the superior performance of our proposed CodingNet algorithm on six datasets with nine baselines.

\begin{figure*}[]
\footnotesize
\begin{minipage}{0.22\linewidth}
\centerline{\includegraphics[width=\textwidth]{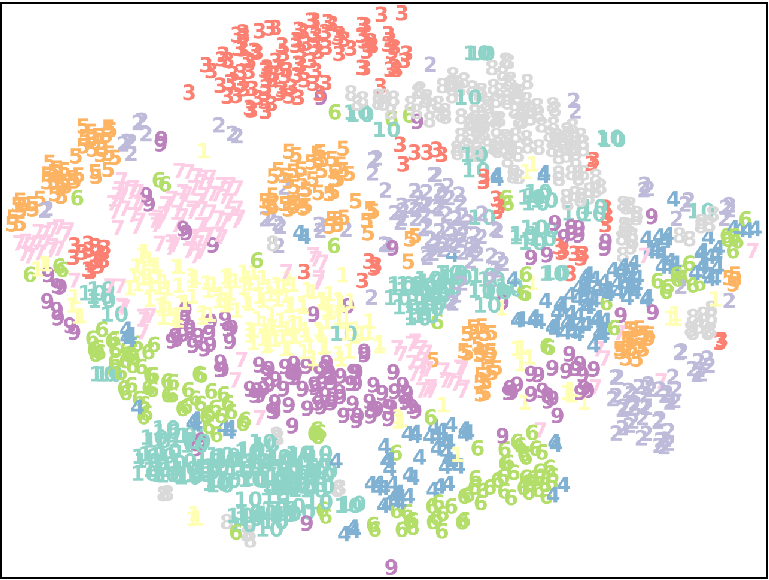}}\hspace{6mm}
\vspace{3pt}
\centerline{Epoch 10}
\centerline{\includegraphics[width=\textwidth]{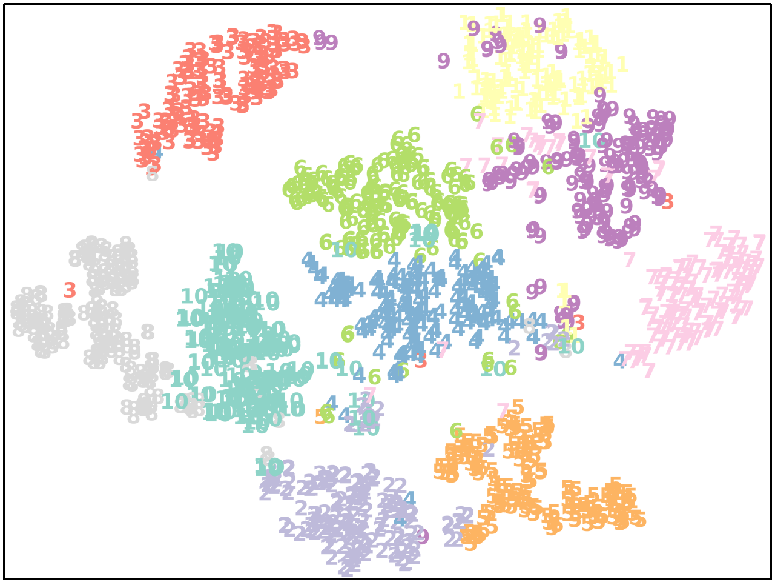}}\hspace{6mm}
\vspace{3pt}
\centerline{Epoch 170}
\end{minipage}
\begin{minipage}{0.22\linewidth}
\centerline{\includegraphics[width=\textwidth]{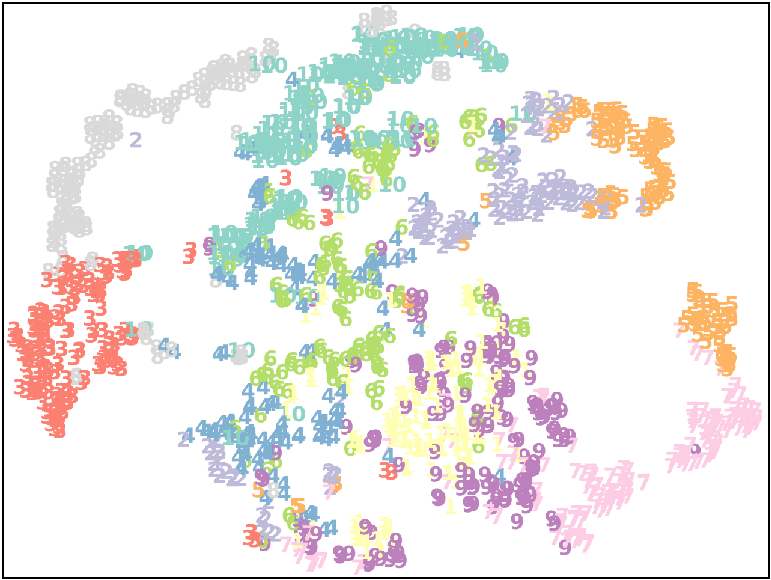}}\hspace{6mm}
\vspace{3pt}
\centerline{Epoch 50}
\centerline{\includegraphics[width=\textwidth]{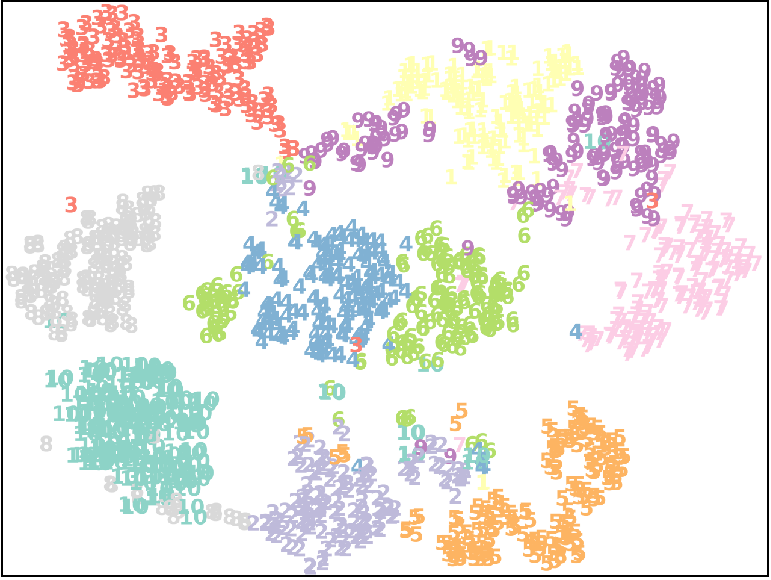}}\hspace{6mm}
\vspace{3pt}
\centerline{Epoch 200}
\end{minipage}
\begin{minipage}{0.22\linewidth}
\centerline{\includegraphics[width=\textwidth]{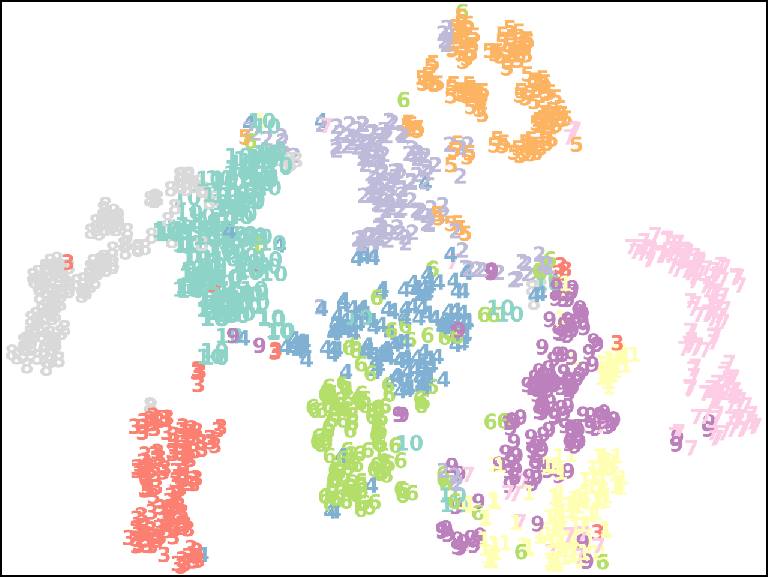}}\hspace{6mm}
\vspace{3pt}
\centerline{Epoch 90}
\centerline{\includegraphics[width=\textwidth]{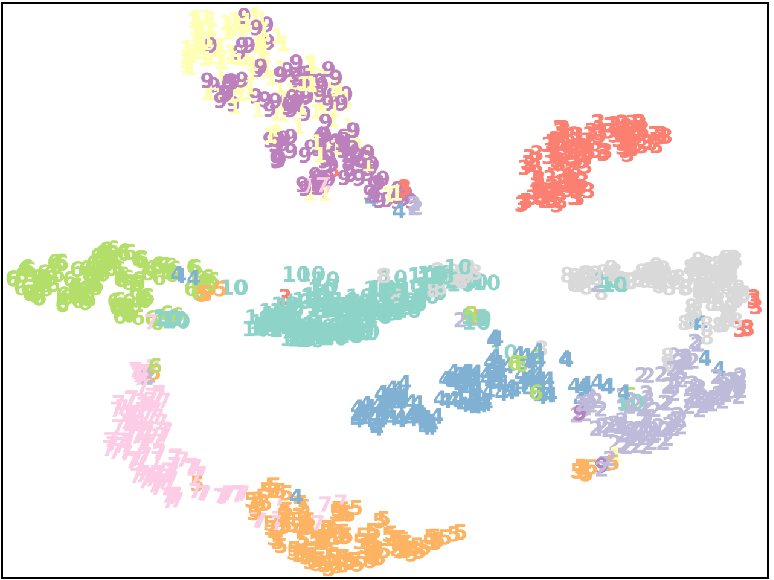}}\hspace{6mm}
\vspace{3pt}
\centerline{Epoch 230}
\end{minipage}
\begin{minipage}{0.22\linewidth}
\centerline{\includegraphics[width=\textwidth]{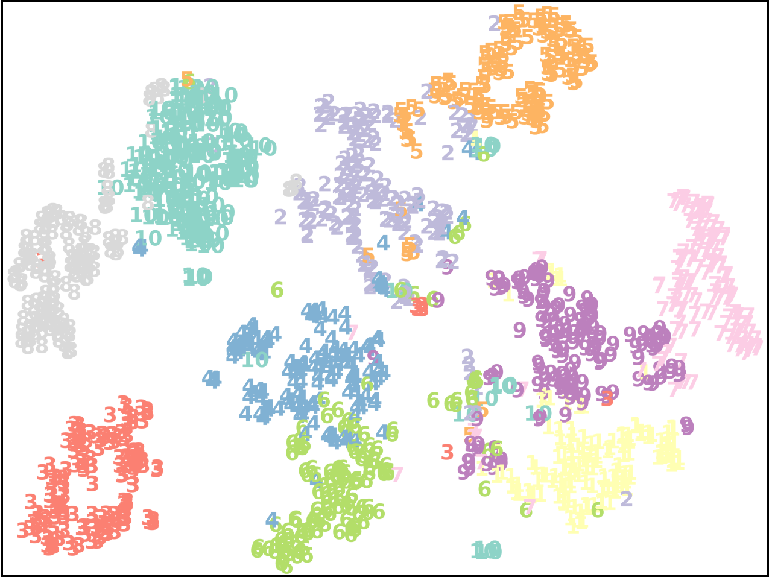}}\hspace{6mm}
\vspace{3pt}
\centerline{Epoch 130}
\centerline{\includegraphics[width=\textwidth]{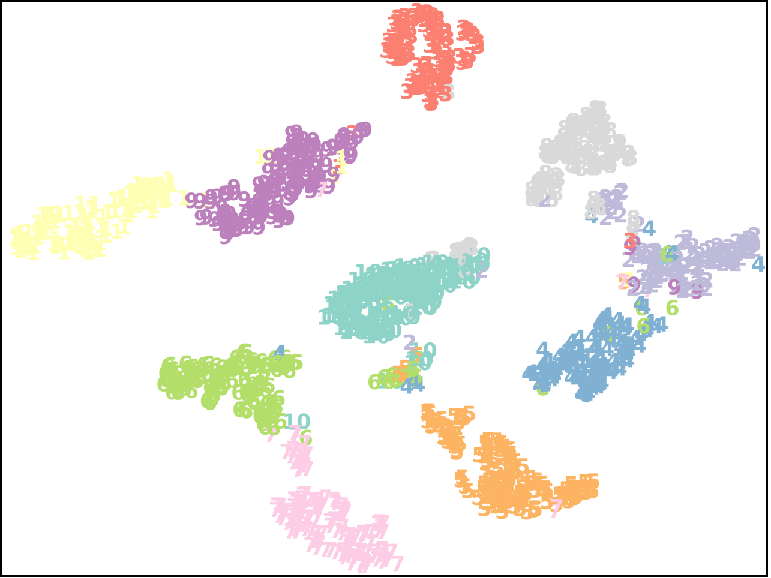}}\hspace{6mm}
\vspace{3pt}
\centerline{Epoch 260}
\end{minipage}
\caption{2D visualization on UCI-digit dataset.}
\label{t_SNE}  
\end{figure*}

\subsection{Ablation Studies~\textbf{(RQ2)}}

\subsubsection{Effectiveness of the designed modules}

In this subsection, we have conducted a series of experiments to validate the efficacy of the proposed strategies, encompassing the consistent learning module, the diverse learning module, and the consistent prediction module. To this end, we employ different model variants denoted as ``(w/o) D'', ``(w/o) C'', ``(w/o) P'', and  ``(w/o) D\&C\&P'' . These variants correspond to reduced models achieved by individually removing the diverse learning module, the consistent learning module, the consistent prediction module, and all aforementioned modules combined. Additionally, we employ an autoencoder as the backbone network to acquire representations for the downstream clustering task, referred to as ``(w/o) D\&C\&P''. The outcomes of these ablation experiments are detailed in Table \ref{ablation_studies}. The overarching conclusion drawn from these results is that the removal of any of our designed modules leads to a noticeable decline in the clustering performance for multi-view data. This empirical evidence underscores the positive contributions of each of the designed modules to the overall clustering performance.

The reasons we analyze are as follows.
\begin{enumerate}
  \item The shallow view features obtained from the diverse learning module capture the fundamental characteristics of multi-view data. By enforcing diversity in these features, the network is guided to generate more distinctive view representations.

  \item The consistency of deep view features is mined by the contrastive loss in both the deep view feature level and pseudo labels level.

  \item Consistency in class information is crucial at the feature level. The consistent prediction module ensures that the network produces the same prediction for both shallow and deep view features, thereby enhancing clustering performance.
\end{enumerate}
In summary, the discriminative capacity of the network can be improved by the proposed modules.

\subsubsection{Effectiveness of the asymmetric network}

We design an asymmetric network to extract both shallow and deep features. In order to emphasize the advantages of this structure, we conducted experiments on six datasets. Specifically, we designed a symmetrical architecture involving four linear layers and three activation functions. The results are presented in Table. \ref{ablation_studies}. It is evident that the model's clustering performance is not promising when employing the symmetrical network. This can be attributed to the following factors:
\begin{enumerate}
  \item The view features extracted through the same encoder network turn out to be quite similar.

  \item Since diverse and consistent aspects of the view features are trained using different loss functions, optimizing the network with disparate loss functions on similar view features becomes challenging. 
\end{enumerate}

\subsubsection{Effectiveness of the diversity and consistency}
In our paper, we the dual goals of enhancing diversity in shallow features and consistency in deep features. Specifically, we enforce the similarity matrix to align with the zero matrix to preserve the diversity of shallow features. Additionally, we introduce a dual contrastive mechanism to ensure the consistency of deep features. In this subsection, we formulate two optimization approaches to validate the impacts of diversity and consistency.  We simultaneously emphasize diversity in both shallow and deep view features, termed as ``\textbf{B-D}''. Similarly, we denote ``\textbf{B-C}'' to indicate an emphasis on consistency for both shallow and deep view features. The experimental findings are presented in Table. \ref{ablation_studies}, revealing the following insights:

\begin{enumerate}
  \item Simultaneously keeping diversity leads to unsatisfactory clustering performance. This result can be attributed to the fact that the fundamental goal of clustering is to group samples into distinct clusters. Overemphasizing diversity between view features runs counter to the clustering objective.

  \item The clustering performance decreases under maintaining the consistency for both shallow and deep view features. This optimization method is equivalent to conducting two levels of contrastive learning, while ignoring the diversity of shallow view features.   
\end{enumerate}

\subsection{Hyper-parameter Analysis~\textbf{(RQ4)}}

In this subsection, we implement experiments to explore the influence of the trade-off parameters in the total loss, i.e., $\mathcal{L}_{rec} + \alpha \mathcal{L}_{div}+\beta \mathcal{L}_{con}+\mathcal{L}_{cp}$. To be concrete, we analyze the influence of the designed diverse learning module and the consistent learning module. The results are conducted in Reuter and Caltech101\_7 with ACC, NMI, and PUR metrics. The trade-off hyper-parameters $\alpha$ and $\beta$ is within the range of $[0.01, 0.1, 1, 10]$. The results are shown in Fig. \ref{Sensitivity}. We could observe as follows. 
\begin{enumerate}
  \item When the value of $\alpha$ and $\beta$ are set to an extreme value, e.g., $0.01$, the model will obtain unpromising clustering performance. We analyze the reason is that the balance of the loss is destroyed. 

  \item $\alpha$ exerts a more substantial impact on model performance. This stems from the fact that the contrastive learning mechanism significantly bolsters the model's discriminative capacity.   
\end{enumerate}

\subsection{Visualization Analysis~\textbf{(RQ5)}}

In this subsection, we employ the $t$-SNE algorithm \cite{T_SNE} to visualize the distribution of learned embeddings from our proposed CodingNet algorithm. We showcase the superiority of CodingNet on the UCI\_digit dataset. The visualization results are depicted in Fig. \ref{t_SNE}, and it is evident that CodingNet effectively revealed the intrinsic clustering structure.

\section{Conclusion}

To concurrently harness diverse and consistent information, this paper introduces CodingNet, an innovative asymmetric double-winged network for multi-view clustering. To be specific, we first propose the asymmetric network to extract the shallow and deep features separately. Besides, two specialized strategies are designed for diversity and consistency. For exploring diversity, we align the similarity matrix on shallow feature to zero matrix, thus reducing the similarity for the shallow features. To explore consistency, we design a dual contrastive mechanism on both view-feature and pseudo-label levels. Extensive experimental results on six datasets demonstrate the effectiveness of CodingNet.

\section*{Acknowledgments}
This work was supported partly by the Pre Research Project (YZ2082300006) from University of Science and Technology of China. 

\bibliographystyle{model1-num-names}

\bibliography{cas-refs}

\begin{thebibliography}{75}
\expandafter\ifx\csname natexlab\endcsname\relax\def\natexlab#1{#1}\fi
\providecommand{\url}[1]{\texttt{#1}}
\providecommand{\href}[2]{#2}
\providecommand{\path}[1]{#1}
\providecommand{\DOIprefix}{doi:}
\providecommand{\ArXivprefix}{arXiv:}
\providecommand{\URLprefix}{URL: }
\providecommand{\Pubmedprefix}{pmid:}
\providecommand{\doi}[1]{\href{http://dx.doi.org/#1}{\path{#1}}}
\providecommand{\Pubmed}[1]{\href{pmid:#1}{\path{#1}}}
\providecommand{\bibinfo}[2]{#2}
\ifx\xfnm\relax \def\xfnm[#1]{\unskip,\space#1}\fi
\bibitem[{Chen et~al.(2020)Chen, Huang, Wang, and Huang}]{chen1}
\bibinfo{author}{M.-S. Chen}, \bibinfo{author}{L.~Huang},
  \bibinfo{author}{C.-D. Wang}, \bibinfo{author}{D.~Huang},
\newblock \bibinfo{title}{Multi-view clustering in latent embedding space},
\newblock in: \bibinfo{booktitle}{Proceedings of the AAAI conference on
  artificial intelligence}, volume~\bibinfo{volume}{34}, \bibinfo{year}{2020},
  pp. \bibinfo{pages}{3513--3520}.
\bibitem[{Chen et~al.(2021)Chen, Huang, Wang, Huang, and Lai}]{chen2}
\bibinfo{author}{M.-S. Chen}, \bibinfo{author}{L.~Huang},
  \bibinfo{author}{C.-D. Wang}, \bibinfo{author}{D.~Huang},
  \bibinfo{author}{J.-H. Lai},
\newblock \bibinfo{title}{Relaxed multi-view clustering in latent embedding
  space},
\newblock \bibinfo{journal}{Information Fusion} \bibinfo{volume}{68}
  (\bibinfo{year}{2021}) \bibinfo{pages}{8--21}.
\bibitem[{Chen et~al.(2022{\natexlab{a}})Chen, Wang, and Lai}]{chen3}
\bibinfo{author}{M.-S. Chen}, \bibinfo{author}{C.-D. Wang},
  \bibinfo{author}{J.-H. Lai},
\newblock \bibinfo{title}{Low-rank tensor based proximity learning for
  multi-view clustering},
\newblock \bibinfo{journal}{IEEE Transactions on Knowledge and Data
  Engineering} \bibinfo{volume}{35} (\bibinfo{year}{2022}{\natexlab{a}})
  \bibinfo{pages}{5076--5090}.
\bibitem[{Chen et~al.(2022{\natexlab{b}})Chen, Liu, Wang, Huang, and
  Lai}]{chen4}
\bibinfo{author}{M.~Chen}, \bibinfo{author}{T.~Liu}, \bibinfo{author}{C.~Wang},
  \bibinfo{author}{D.~Huang}, \bibinfo{author}{J.~Lai},
\newblock \bibinfo{title}{Adaptively-weighted integral space for fast multiview
  clustering},
\newblock in: \bibinfo{booktitle}{The 30th {ACM} International Conference on
  Multimedia}, \bibinfo{publisher}{{ACM}}, \bibinfo{year}{2022}{\natexlab{b}},
  pp. \bibinfo{pages}{3774--3782}.
\bibitem[{Yang et~al.(2023)Yang, Jin, Wang, Liang, Liu, Wen, Liu, Zhou, Liu,
  and Zhu}]{DealMVC}
\bibinfo{author}{X.~Yang}, \bibinfo{author}{J.~Jin}, \bibinfo{author}{S.~Wang},
  \bibinfo{author}{K.~Liang}, \bibinfo{author}{Y.~Liu},
  \bibinfo{author}{Y.~Wen}, \bibinfo{author}{S.~Liu},
  \bibinfo{author}{S.~Zhou}, \bibinfo{author}{X.~Liu},
  \bibinfo{author}{E.~Zhu},
\newblock \bibinfo{title}{Dealmvc: Dual contrastive calibration for multi-view
  clustering},
\newblock in: \bibinfo{booktitle}{Proceedings of the 31th ACM International
  Conference on Multimedia}, \bibinfo{year}{2023}.
\bibitem[{Liu et~al.(2022)Liu, Liu, Wang, Niu, and Zhu}]{suyuan_TNNLS}
\bibinfo{author}{S.~Liu}, \bibinfo{author}{X.~Liu}, \bibinfo{author}{S.~Wang},
  \bibinfo{author}{X.~Niu}, \bibinfo{author}{E.~Zhu},
\newblock \bibinfo{title}{Fast incomplete multi-view clustering with
  view-independent anchors},
\newblock \bibinfo{journal}{IEEE Transactions on Neural Networks and Learning
  Systems}  (\bibinfo{year}{2022}).
\bibitem[{Wen et~al.(2018)Wen, Zhang, Xu, and Zhong}]{wenjie_mf}
\bibinfo{author}{J.~Wen}, \bibinfo{author}{Z.~Zhang}, \bibinfo{author}{Y.~Xu},
  \bibinfo{author}{Z.~Zhong},
\newblock \bibinfo{title}{Incomplete multi-view clustering via graph
  regularized matrix factorization},
\newblock in: \bibinfo{booktitle}{Proceedings of the European conference on
  computer vision (ECCV) workshops}, \bibinfo{year}{2018}, pp.
  \bibinfo{pages}{0--0}.
\bibitem[{Liu et~al.(2022)Liu, Wang, Zhang, Xu, Liu, Zhang, and
  Gao}]{suyuan_AAAI}
\bibinfo{author}{S.~Liu}, \bibinfo{author}{S.~Wang},
  \bibinfo{author}{P.~Zhang}, \bibinfo{author}{K.~Xu},
  \bibinfo{author}{X.~Liu}, \bibinfo{author}{C.~Zhang},
  \bibinfo{author}{F.~Gao},
\newblock \bibinfo{title}{Efficient one-pass multi-view subspace clustering
  with consensus anchors},
\newblock in: \bibinfo{booktitle}{Proceedings of the AAAI Conference on
  Artificial Intelligence}, volume~\bibinfo{volume}{36}, \bibinfo{year}{2022},
  pp. \bibinfo{pages}{7576--7584}.
\bibitem[{Wang et~al.(2022)Wang, Liu, Liu, Jin, Tu, Zhu, and Zhu}]{siwei_nips}
\bibinfo{author}{S.~Wang}, \bibinfo{author}{X.~Liu}, \bibinfo{author}{S.~Liu},
  \bibinfo{author}{J.~Jin}, \bibinfo{author}{W.~Tu}, \bibinfo{author}{X.~Zhu},
  \bibinfo{author}{E.~Zhu},
\newblock \bibinfo{title}{Align then fusion: Generalized large-scale multi-view
  clustering with anchor matching correspondences},
\newblock \bibinfo{journal}{Advances in Neural Information Processing Systems}
  \bibinfo{volume}{35} (\bibinfo{year}{2022}) \bibinfo{pages}{5882--5895}.
\bibitem[{Liu et~al.(2021)Liu, Liu, Liao, Wang, Zhang, Tu, Tang, Liu, and
  Zhu}]{one}
\bibinfo{author}{X.~Liu}, \bibinfo{author}{L.~Liu}, \bibinfo{author}{Q.~Liao},
  \bibinfo{author}{S.~Wang}, \bibinfo{author}{Y.~Zhang},
  \bibinfo{author}{W.~Tu}, \bibinfo{author}{C.~Tang}, \bibinfo{author}{J.~Liu},
  \bibinfo{author}{E.~Zhu},
\newblock \bibinfo{title}{One pass late fusion multi-view clustering},
\newblock in: \bibinfo{booktitle}{International Conference on Machine
  Learning}, \bibinfo{organization}{PMLR}, \bibinfo{year}{2021}, pp.
  \bibinfo{pages}{6850--6859}.
\bibitem[{Liu(2022)}]{simplemkkm}
\bibinfo{author}{X.~Liu},
\newblock \bibinfo{title}{Simplemkkm: Simple multiple kernel k-means},
\newblock \bibinfo{journal}{IEEE Transactions on Pattern Analysis and Machine
  Intelligence}  (\bibinfo{year}{2022}).
\bibitem[{Liu(2021)}]{incomplete}
\bibinfo{author}{X.~Liu},
\newblock \bibinfo{title}{Incomplete multiple kernel alignment maximization for
  clustering},
\newblock \bibinfo{journal}{IEEE Transactions on Pattern Analysis and Machine
  Intelligence}  (\bibinfo{year}{2021}).
\bibitem[{Liu et~al.(2020)Liu, Li, Tang, Xia, Xiong, Liu, Kloft, and
  Zhu}]{efficient}
\bibinfo{author}{X.~Liu}, \bibinfo{author}{M.~Li}, \bibinfo{author}{C.~Tang},
  \bibinfo{author}{J.~Xia}, \bibinfo{author}{J.~Xiong},
  \bibinfo{author}{L.~Liu}, \bibinfo{author}{M.~Kloft},
  \bibinfo{author}{E.~Zhu},
\newblock \bibinfo{title}{Efficient and effective regularized incomplete
  multi-view clustering},
\newblock \bibinfo{journal}{IEEE transactions on pattern analysis and machine
  intelligence} \bibinfo{volume}{43} (\bibinfo{year}{2020})
  \bibinfo{pages}{2634--2646}.
\bibitem[{Zhou et~al.(2019)Zhou, Liu, Li, Zhu, Liu, Zhang, and Yin}]{ZHOU_1}
\bibinfo{author}{S.~Zhou}, \bibinfo{author}{X.~Liu}, \bibinfo{author}{M.~Li},
  \bibinfo{author}{E.~Zhu}, \bibinfo{author}{L.~Liu},
  \bibinfo{author}{C.~Zhang}, \bibinfo{author}{J.~Yin},
\newblock \bibinfo{title}{Multiple kernel clustering with neighbor-kernel
  subspace segmentation},
\newblock \bibinfo{journal}{IEEE transactions on neural networks and learning
  systems} \bibinfo{volume}{31} (\bibinfo{year}{2019})
  \bibinfo{pages}{1351--1362}.
\bibitem[{Zhou et~al.(2020)Zhou, Zhu, Liu, Zheng, Liu, Xia, and Yin}]{ZHOU_2}
\bibinfo{author}{S.~Zhou}, \bibinfo{author}{E.~Zhu}, \bibinfo{author}{X.~Liu},
  \bibinfo{author}{T.~Zheng}, \bibinfo{author}{Q.~Liu},
  \bibinfo{author}{J.~Xia}, \bibinfo{author}{J.~Yin},
\newblock \bibinfo{title}{Subspace segmentation-based robust multiple kernel
  clustering},
\newblock \bibinfo{journal}{Information Fusion} \bibinfo{volume}{53}
  (\bibinfo{year}{2020}) \bibinfo{pages}{145--154}.
\bibitem[{Zhang et~al.(2020)Zhang, Wang, Hu, Cheng, Guo, Zhu, and
  Cai}]{zhang2020adaptive}
\bibinfo{author}{P.~Zhang}, \bibinfo{author}{S.~Wang}, \bibinfo{author}{J.~Hu},
  \bibinfo{author}{Z.~Cheng}, \bibinfo{author}{X.~Guo},
  \bibinfo{author}{E.~Zhu}, \bibinfo{author}{Z.~Cai},
\newblock \bibinfo{title}{Adaptive weighted graph fusion incomplete multi-view
  subspace clustering},
\newblock \bibinfo{journal}{Sensors} \bibinfo{volume}{20}
  (\bibinfo{year}{2020}) \bibinfo{pages}{5755}.
\bibitem[{Xu et~al.(2022)Xu, Tang, Ren, Peng, Zhu, and He}]{MFLVC}
\bibinfo{author}{J.~Xu}, \bibinfo{author}{H.~Tang}, \bibinfo{author}{Y.~Ren},
  \bibinfo{author}{L.~Peng}, \bibinfo{author}{X.~Zhu}, \bibinfo{author}{L.~He},
\newblock \bibinfo{title}{Multi-level feature learning for contrastive
  multi-view clustering},
\newblock in: \bibinfo{booktitle}{Proceedings of the IEEE/CVF Conference on
  Computer Vision and Pattern Recognition}, \bibinfo{year}{2022}, pp.
  \bibinfo{pages}{16051--16060}.
\bibitem[{Liu et~al.(2022)Liu, Liao, Ma, and Zhan}]{SDSNE}
\bibinfo{author}{C.~Liu}, \bibinfo{author}{Z.~Liao}, \bibinfo{author}{Y.~Ma},
  \bibinfo{author}{K.~Zhan},
\newblock \bibinfo{title}{Stationary diffusion state neural estimation for
  multiview clustering},
\newblock in: \bibinfo{booktitle}{Proceedings of the AAAI Conference on
  Artificial Intelligence}, volume~\bibinfo{volume}{36}, \bibinfo{year}{2022},
  pp. \bibinfo{pages}{7542--7549}.
\bibitem[{Peng et~al.(2019)Peng, Huang, Lv, Zhu, and Zhou}]{COMIC}
\bibinfo{author}{X.~Peng}, \bibinfo{author}{Z.~Huang}, \bibinfo{author}{J.~Lv},
  \bibinfo{author}{H.~Zhu}, \bibinfo{author}{J.~T. Zhou},
\newblock \bibinfo{title}{Comic: Multi-view clustering without parameter
  selection},
\newblock in: \bibinfo{booktitle}{International conference on machine
  learning}, \bibinfo{organization}{PMLR}, \bibinfo{year}{2019}, pp.
  \bibinfo{pages}{5092--5101}.
\bibitem[{Xu et~al.(2022)Xu, Ren, Tang, Yang, Pan, Yang, Pu, Philip, and
  He}]{SDMVC}
\bibinfo{author}{J.~Xu}, \bibinfo{author}{Y.~Ren}, \bibinfo{author}{H.~Tang},
  \bibinfo{author}{Z.~Yang}, \bibinfo{author}{L.~Pan},
  \bibinfo{author}{Y.~Yang}, \bibinfo{author}{X.~Pu}, \bibinfo{author}{S.~Y.
  Philip}, \bibinfo{author}{L.~He},
\newblock \bibinfo{title}{Self-supervised discriminative feature learning for
  deep multi-view clustering},
\newblock \bibinfo{journal}{IEEE Transactions on Knowledge and Data
  Engineering}  (\bibinfo{year}{2022}).
\bibitem[{Lin et~al.(2021)Lin, Gou, Liu, Li, Lv, and Peng}]{dmvc1}
\bibinfo{author}{Y.~Lin}, \bibinfo{author}{Y.~Gou}, \bibinfo{author}{Z.~Liu},
  \bibinfo{author}{B.~Li}, \bibinfo{author}{J.~Lv}, \bibinfo{author}{X.~Peng},
\newblock \bibinfo{title}{Completer: Incomplete multi-view clustering via
  contrastive prediction},
\newblock in: \bibinfo{booktitle}{Proceedings of the IEEE/CVF conference on
  computer vision and pattern recognition}, \bibinfo{year}{2021}, pp.
  \bibinfo{pages}{11174--11183}.
\bibitem[{Xu et~al.(2022)Xu, Ren, Tang, Yang, Pan, Yang, Pu, Philip, and
  He}]{dmvc3}
\bibinfo{author}{J.~Xu}, \bibinfo{author}{Y.~Ren}, \bibinfo{author}{H.~Tang},
  \bibinfo{author}{Z.~Yang}, \bibinfo{author}{L.~Pan},
  \bibinfo{author}{Y.~Yang}, \bibinfo{author}{X.~Pu}, \bibinfo{author}{S.~Y.
  Philip}, \bibinfo{author}{L.~He},
\newblock \bibinfo{title}{Self-supervised discriminative feature learning for
  deep multi-view clustering},
\newblock \bibinfo{journal}{IEEE Transactions on Knowledge and Data
  Engineering}  (\bibinfo{year}{2022}).
\bibitem[{Xu et~al.(2021)Xu, Ren, Li, Pan, Zhu, and Xu}]{dmvc4}
\bibinfo{author}{J.~Xu}, \bibinfo{author}{Y.~Ren}, \bibinfo{author}{G.~Li},
  \bibinfo{author}{L.~Pan}, \bibinfo{author}{C.~Zhu}, \bibinfo{author}{Z.~Xu},
\newblock \bibinfo{title}{Deep embedded multi-view clustering with
  collaborative training},
\newblock \bibinfo{journal}{Information Sciences} \bibinfo{volume}{573}
  (\bibinfo{year}{2021}) \bibinfo{pages}{279--290}.
\bibitem[{Lin et~al.(2021)Lin, Gou, Liu, Li, Lv, and Peng}]{Completer}
\bibinfo{author}{Y.~Lin}, \bibinfo{author}{Y.~Gou}, \bibinfo{author}{Z.~Liu},
  \bibinfo{author}{B.~Li}, \bibinfo{author}{J.~Lv}, \bibinfo{author}{X.~Peng},
\newblock \bibinfo{title}{Completer: Incomplete multi-view clustering via
  contrastive prediction},
\newblock in: \bibinfo{booktitle}{Proceedings of the IEEE/CVF conference on
  computer vision and pattern recognition}, \bibinfo{year}{2021}, pp.
  \bibinfo{pages}{11174--11183}.
\bibitem[{Zhou and Shen(2020)}]{EAMC}
\bibinfo{author}{R.~Zhou}, \bibinfo{author}{Y.-D. Shen},
\newblock \bibinfo{title}{End-to-end adversarial-attention network for
  multi-modal clustering},
\newblock in: \bibinfo{booktitle}{Proceedings of the IEEE/CVF conference on
  computer vision and pattern recognition}, \bibinfo{year}{2020}, pp.
  \bibinfo{pages}{14619--14628}.
\bibitem[{Trosten et~al.(2021)Trosten, Lokse, Jenssen, and Kampffmeyer}]{SiMVC}
\bibinfo{author}{D.~J. Trosten}, \bibinfo{author}{S.~Lokse},
  \bibinfo{author}{R.~Jenssen}, \bibinfo{author}{M.~Kampffmeyer},
\newblock \bibinfo{title}{Reconsidering representation alignment for multi-view
  clustering},
\newblock in: \bibinfo{booktitle}{Proceedings of the IEEE/CVF conference on
  computer vision and pattern recognition}, \bibinfo{year}{2021}, pp.
  \bibinfo{pages}{1255--1265}.
\bibitem[{Cheng et~al.(2021)Cheng, Wang, Tao, Xie, and Gao}]{graph_m1}
\bibinfo{author}{J.~Cheng}, \bibinfo{author}{Q.~Wang},
  \bibinfo{author}{Z.~Tao}, \bibinfo{author}{D.~Xie}, \bibinfo{author}{Q.~Gao},
\newblock \bibinfo{title}{Multi-view attribute graph convolution networks for
  clustering},
\newblock in: \bibinfo{booktitle}{Proceedings of the twenty-ninth international
  conference on international joint conferences on artificial intelligence},
  \bibinfo{year}{2021}, pp. \bibinfo{pages}{2973--2979}.
\bibitem[{Huang et~al.(2020)Huang, Kang, and Xu}]{graph_m2}
\bibinfo{author}{S.~Huang}, \bibinfo{author}{Z.~Kang}, \bibinfo{author}{Z.~Xu},
\newblock \bibinfo{title}{Auto-weighted multi-view clustering via deep matrix
  decomposition},
\newblock \bibinfo{journal}{Pattern Recognition} \bibinfo{volume}{97}
  (\bibinfo{year}{2020}) \bibinfo{pages}{107015}.
\bibitem[{Li et~al.(2019)Li, Wang, Tao, Gao, Yang et~al.}]{adver_m1}
\bibinfo{author}{Z.~Li}, \bibinfo{author}{Q.~Wang}, \bibinfo{author}{Z.~Tao},
  \bibinfo{author}{Q.~Gao}, \bibinfo{author}{Z.~Yang}, et~al.,
\newblock \bibinfo{title}{Deep adversarial multi-view clustering network.},
\newblock in: \bibinfo{booktitle}{IJCAI}, \bibinfo{year}{2019}, pp.
  \bibinfo{pages}{2952--2958}.
\bibitem[{Zhou and Shen(2020)}]{adver_m2}
\bibinfo{author}{R.~Zhou}, \bibinfo{author}{Y.-D. Shen},
\newblock \bibinfo{title}{End-to-end adversarial-attention network for
  multi-modal clustering},
\newblock in: \bibinfo{booktitle}{Proceedings of the IEEE/CVF conference on
  computer vision and pattern recognition}, \bibinfo{year}{2020}, pp.
  \bibinfo{pages}{14619--14628}.
\bibitem[{Liu et~al.(2013)Liu, Wang, Gao, and Han}]{mvc_mf}
\bibinfo{author}{J.~Liu}, \bibinfo{author}{C.~Wang}, \bibinfo{author}{J.~Gao},
  \bibinfo{author}{J.~Han},
\newblock \bibinfo{title}{Multi-view clustering via joint nonnegative matrix
  factorization},
\newblock in: \bibinfo{booktitle}{Proceedings of the 2013 SIAM international
  conference on data mining}, \bibinfo{organization}{SIAM},
  \bibinfo{year}{2013}, pp. \bibinfo{pages}{252--260}.
\bibitem[{Zhao et~al.(2017)Zhao, Ding, and Fu}]{mvc_mf2}
\bibinfo{author}{H.~Zhao}, \bibinfo{author}{Z.~Ding}, \bibinfo{author}{Y.~Fu},
\newblock \bibinfo{title}{Multi-view clustering via deep matrix factorization},
\newblock in: \bibinfo{booktitle}{Proceedings of the AAAI conference on
  artificial intelligence}, volume~\bibinfo{volume}{31}, \bibinfo{year}{2017}.
\bibitem[{Cai et~al.(2013)Cai, Nie, and Huang}]{mvc_mf3}
\bibinfo{author}{X.~Cai}, \bibinfo{author}{F.~Nie}, \bibinfo{author}{H.~Huang},
\newblock \bibinfo{title}{Multi-view k-means clustering on big data},
\newblock in: \bibinfo{booktitle}{Twenty-Third International Joint conference
  on artificial intelligence}, \bibinfo{year}{2013}.
\bibitem[{Tian et~al.(2020)Tian, Krishnan, and Isola}]{CMC}
\bibinfo{author}{Y.~Tian}, \bibinfo{author}{D.~Krishnan},
  \bibinfo{author}{P.~Isola},
\newblock \bibinfo{title}{Contrastive multiview coding},
\newblock in: \bibinfo{booktitle}{Computer Vision--ECCV 2020: 16th European
  Conference, Glasgow, UK, August 23--28, 2020, Proceedings, Part XI 16},
  \bibinfo{organization}{Springer}, \bibinfo{year}{2020}, pp.
  \bibinfo{pages}{776--794}.
\bibitem[{Li et~al.(2021)Li, Tang, Liu, Zheng, Zhang, and Zhu}]{graph_mvc}
\bibinfo{author}{Z.~Li}, \bibinfo{author}{C.~Tang}, \bibinfo{author}{X.~Liu},
  \bibinfo{author}{X.~Zheng}, \bibinfo{author}{W.~Zhang},
  \bibinfo{author}{E.~Zhu},
\newblock \bibinfo{title}{Consensus graph learning for multi-view clustering},
\newblock \bibinfo{journal}{IEEE Transactions on Multimedia}
  \bibinfo{volume}{24} (\bibinfo{year}{2021}) \bibinfo{pages}{2461--2472}.
\bibitem[{Liu et~al.(2022)Liu, Chen, Luo, Kang, Luo, and Han}]{kang_graph}
\bibinfo{author}{L.~Liu}, \bibinfo{author}{P.~Chen}, \bibinfo{author}{G.~Luo},
  \bibinfo{author}{Z.~Kang}, \bibinfo{author}{Y.~Luo},
  \bibinfo{author}{S.~Han},
\newblock \bibinfo{title}{Scalable multi-view clustering with graph filtering},
\newblock \bibinfo{journal}{Neural Computing and Applications}
  \bibinfo{volume}{34} (\bibinfo{year}{2022}) \bibinfo{pages}{16213--16221}.
\bibitem[{Xia et~al.(2021)Xia, Wang, Gao, Zhang, and Gao}]{xia_graph}
\bibinfo{author}{W.~Xia}, \bibinfo{author}{Q.~Wang}, \bibinfo{author}{Q.~Gao},
  \bibinfo{author}{X.~Zhang}, \bibinfo{author}{X.~Gao},
\newblock \bibinfo{title}{Self-supervised graph convolutional network for
  multi-view clustering},
\newblock \bibinfo{journal}{IEEE Transactions on Multimedia}
  \bibinfo{volume}{24} (\bibinfo{year}{2021}) \bibinfo{pages}{3182--3192}.
\bibitem[{Xia et~al.(2022{\natexlab{a}})Xia, Gao, Wang, Gao, Ding, and
  Tao}]{xia_graph1}
\bibinfo{author}{W.~Xia}, \bibinfo{author}{Q.~Gao}, \bibinfo{author}{Q.~Wang},
  \bibinfo{author}{X.~Gao}, \bibinfo{author}{C.~Ding},
  \bibinfo{author}{D.~Tao},
\newblock \bibinfo{title}{Tensorized bipartite graph learning for multi-view
  clustering},
\newblock \bibinfo{journal}{IEEE Transactions on Pattern Analysis and Machine
  Intelligence} \bibinfo{volume}{45} (\bibinfo{year}{2022}{\natexlab{a}})
  \bibinfo{pages}{5187--5202}.
\bibitem[{Xia et~al.(2022{\natexlab{b}})Xia, Wang, Yang, Gao, Han, and
  Gao}]{xia_graph3}
\bibinfo{author}{W.~Xia}, \bibinfo{author}{S.~Wang}, \bibinfo{author}{M.~Yang},
  \bibinfo{author}{Q.~Gao}, \bibinfo{author}{J.~Han}, \bibinfo{author}{X.~Gao},
\newblock \bibinfo{title}{Multi-view graph embedding clustering network: Joint
  self-supervision and block diagonal representation},
\newblock \bibinfo{journal}{Neural Networks} \bibinfo{volume}{145}
  (\bibinfo{year}{2022}{\natexlab{b}}) \bibinfo{pages}{1--9}.
\bibitem[{Kang et~al.(2020)Kang, Zhou, Zhao, Shao, Han, and Xu}]{kang_subspace}
\bibinfo{author}{Z.~Kang}, \bibinfo{author}{W.~Zhou},
  \bibinfo{author}{Z.~Zhao}, \bibinfo{author}{J.~Shao},
  \bibinfo{author}{M.~Han}, \bibinfo{author}{Z.~Xu},
\newblock \bibinfo{title}{Large-scale multi-view subspace clustering in linear
  time},
\newblock in: \bibinfo{booktitle}{Proceedings of the AAAI conference on
  artificial intelligence}, volume~\bibinfo{volume}{34}, \bibinfo{year}{2020},
  pp. \bibinfo{pages}{4412--4419}.
\bibitem[{Gao et~al.(2020)Gao, Xia, Wan, Xie, and Zhang}]{xia_sub}
\bibinfo{author}{Q.~Gao}, \bibinfo{author}{W.~Xia}, \bibinfo{author}{Z.~Wan},
  \bibinfo{author}{D.~Xie}, \bibinfo{author}{P.~Zhang},
\newblock \bibinfo{title}{Tensor-svd based graph learning for multi-view
  subspace clustering},
\newblock in: \bibinfo{booktitle}{Proceedings of the AAAI Conference on
  Artificial Intelligence}, volume~\bibinfo{volume}{34}, \bibinfo{year}{2020},
  pp. \bibinfo{pages}{3930--3937}.
\bibitem[{Cao et~al.(2015)Cao, Zhang, Fu, Liu, and Zhang}]{subspace}
\bibinfo{author}{X.~Cao}, \bibinfo{author}{C.~Zhang}, \bibinfo{author}{H.~Fu},
  \bibinfo{author}{S.~Liu}, \bibinfo{author}{H.~Zhang},
\newblock \bibinfo{title}{Diversity-induced multi-view subspace clustering},
\newblock in: \bibinfo{booktitle}{Proceedings of the IEEE conference on
  computer vision and pattern recognition}, \bibinfo{year}{2015}, pp.
  \bibinfo{pages}{586--594}.
\bibitem[{Zhang and Wu(2012)}]{zhang1}
\bibinfo{author}{X.-L. Zhang}, \bibinfo{author}{J.~Wu},
\newblock \bibinfo{title}{Deep belief networks based voice activity detection},
\newblock \bibinfo{journal}{IEEE Transactions on Audio, Speech, and Language
  Processing} \bibinfo{volume}{21} (\bibinfo{year}{2012})
  \bibinfo{pages}{697--710}.
\bibitem[{Zhang and Wang(2016)}]{zhang2}
\bibinfo{author}{X.-L. Zhang}, \bibinfo{author}{D.~Wang},
\newblock \bibinfo{title}{A deep ensemble learning method for monaural speech
  separation},
\newblock \bibinfo{journal}{IEEE/ACM transactions on audio, speech, and
  language processing} \bibinfo{volume}{24} (\bibinfo{year}{2016})
  \bibinfo{pages}{967--977}.
\bibitem[{Zhang and Wang(2015)}]{zhang3}
\bibinfo{author}{X.-L. Zhang}, \bibinfo{author}{D.~Wang},
\newblock \bibinfo{title}{Boosting contextual information for deep neural
  network based voice activity detection},
\newblock \bibinfo{journal}{IEEE/ACM Transactions on Audio, Speech, and
  Language Processing} \bibinfo{volume}{24} (\bibinfo{year}{2015})
  \bibinfo{pages}{252--264}.
\bibitem[{Zhang and Wang(2014)}]{zhang4}
\bibinfo{author}{X.-L. Zhang}, \bibinfo{author}{D.~Wang},
\newblock \bibinfo{title}{Boosted deep neural networks and multi-resolution
  cochleagram features for voice activity detection},
\newblock in: \bibinfo{booktitle}{Fifteenth annual conference of the
  international speech communication association}, \bibinfo{year}{2014}.
\bibitem[{Yang et~al.(2022)Yang, Liu, Zhou, Liu, and Zhu}]{MGCN}
\bibinfo{author}{X.~Yang}, \bibinfo{author}{Y.~Liu}, \bibinfo{author}{S.~Zhou},
  \bibinfo{author}{X.~Liu}, \bibinfo{author}{E.~Zhu},
\newblock \bibinfo{title}{Mixed graph contrastive network for semi-supervised
  node classification},
\newblock \bibinfo{journal}{arXiv preprint arXiv:2206.02796}
  (\bibinfo{year}{2022}).
\bibitem[{Li et~al.(2019)Li, Wang, Tao, Gao, Yang et~al.}]{DAMC}
\bibinfo{author}{Z.~Li}, \bibinfo{author}{Q.~Wang}, \bibinfo{author}{Z.~Tao},
  \bibinfo{author}{Q.~Gao}, \bibinfo{author}{Z.~Yang}, et~al.,
\newblock \bibinfo{title}{Deep adversarial multi-view clustering network.},
\newblock in: \bibinfo{booktitle}{IJCAI}, \bibinfo{year}{2019}, pp.
  \bibinfo{pages}{2952--2958}.
\bibitem[{Xu et~al.(2021)Xu, Ren, Tang, Pu, Zhu, Zeng, and He}]{MultiVAE}
\bibinfo{author}{J.~Xu}, \bibinfo{author}{Y.~Ren}, \bibinfo{author}{H.~Tang},
  \bibinfo{author}{X.~Pu}, \bibinfo{author}{X.~Zhu}, \bibinfo{author}{M.~Zeng},
  \bibinfo{author}{L.~He},
\newblock \bibinfo{title}{Multi-vae: Learning disentangled view-common and
  view-peculiar visual representations for multi-view clustering},
\newblock in: \bibinfo{booktitle}{Proceedings of the IEEE/CVF International
  Conference on Computer Vision}, \bibinfo{year}{2021}, pp.
  \bibinfo{pages}{9234--9243}.
\bibitem[{Hjelm et~al.(2018)Hjelm, Fedorov, Lavoie-Marchildon, Grewal, Bachman,
  Trischler, and Bengio}]{DIM}
\bibinfo{author}{R.~D. Hjelm}, \bibinfo{author}{A.~Fedorov},
  \bibinfo{author}{S.~Lavoie-Marchildon}, \bibinfo{author}{K.~Grewal},
  \bibinfo{author}{P.~Bachman}, \bibinfo{author}{A.~Trischler},
  \bibinfo{author}{Y.~Bengio},
\newblock \bibinfo{title}{Learning deep representations by mutual information
  estimation and maximization},
\newblock in: \bibinfo{booktitle}{International Conference on Learning
  Representations}, \bibinfo{year}{2018}.
\bibitem[{Chen et~al.(2020)Chen, Kornblith, Norouzi, and Hinton}]{SIMCLR}
\bibinfo{author}{T.~Chen}, \bibinfo{author}{S.~Kornblith},
  \bibinfo{author}{M.~Norouzi}, \bibinfo{author}{G.~Hinton},
\newblock \bibinfo{title}{A simple framework for contrastive learning of visual
  representations},
\newblock in: \bibinfo{booktitle}{International conference on machine
  learning}, \bibinfo{organization}{PMLR}, \bibinfo{year}{2020}, pp.
  \bibinfo{pages}{1597--1607}.
\bibitem[{Grill et~al.(2020)Grill, Strub, Altch{\'e}, Tallec, Richemond,
  Buchatskaya, Doersch, Pires, Guo, Azar et~al.}]{BYOL}
\bibinfo{author}{J.-B. Grill}, \bibinfo{author}{F.~Strub},
  \bibinfo{author}{F.~Altch{\'e}}, \bibinfo{author}{C.~Tallec},
  \bibinfo{author}{P.~H. Richemond}, \bibinfo{author}{E.~Buchatskaya},
  \bibinfo{author}{C.~Doersch}, \bibinfo{author}{B.~A. Pires},
  \bibinfo{author}{Z.~D. Guo}, \bibinfo{author}{M.~G. Azar}, et~al.,
\newblock \bibinfo{title}{Bootstrap your own latent: A new approach to
  self-supervised learning},
\newblock \bibinfo{journal}{arXiv preprint arXiv:2006.07733}
  (\bibinfo{year}{2020}).
\bibitem[{Zbontar et~al.(2021)Zbontar, Jing, Misra, LeCun, and Deny}]{BARLOW}
\bibinfo{author}{J.~Zbontar}, \bibinfo{author}{L.~Jing},
  \bibinfo{author}{I.~Misra}, \bibinfo{author}{Y.~LeCun},
  \bibinfo{author}{S.~Deny},
\newblock \bibinfo{title}{Barlow twins: Self-supervised learning via redundancy
  reduction},
\newblock \bibinfo{journal}{arXiv preprint arXiv:2103.03230}
  (\bibinfo{year}{2021}).
\bibitem[{Zhong et~al.(2021)Zhong, Wu, Chen, Huang, Deng, Nie, Lin, and
  Hua}]{GCC}
\bibinfo{author}{H.~Zhong}, \bibinfo{author}{J.~Wu}, \bibinfo{author}{C.~Chen},
  \bibinfo{author}{J.~Huang}, \bibinfo{author}{M.~Deng},
  \bibinfo{author}{L.~Nie}, \bibinfo{author}{Z.~Lin}, \bibinfo{author}{X.-S.
  Hua},
\newblock \bibinfo{title}{Graph contrastive clustering},
\newblock in: \bibinfo{booktitle}{Proceedings of the IEEE/CVF International
  Conference on Computer Vision}, \bibinfo{year}{2021}, pp.
  \bibinfo{pages}{9224--9233}.
\bibitem[{Yang et~al.(2022)Yang, Hu, Zhou, Liu, and Zhu}]{xihong}
\bibinfo{author}{X.~Yang}, \bibinfo{author}{X.~Hu}, \bibinfo{author}{S.~Zhou},
  \bibinfo{author}{X.~Liu}, \bibinfo{author}{E.~Zhu},
\newblock \bibinfo{title}{Interpolation-based contrastive learning for
  few-label semi-supervised learning},
\newblock \bibinfo{journal}{IEEE Transactions on Neural Networks and Learning
  Systems}  (\bibinfo{year}{2022}) \bibinfo{pages}{1--12}.
\bibitem[{Liu et~al.(2022)Liu, Tu, Zhou, Liu, Song, Yang, and Zhu}]{DCRN}
\bibinfo{author}{Y.~Liu}, \bibinfo{author}{W.~Tu}, \bibinfo{author}{S.~Zhou},
  \bibinfo{author}{X.~Liu}, \bibinfo{author}{L.~Song},
  \bibinfo{author}{X.~Yang}, \bibinfo{author}{E.~Zhu},
\newblock \bibinfo{title}{Deep graph clustering via dual correlation
  reduction},
\newblock in: \bibinfo{booktitle}{AAAI Conference on Artificial Intelligence},
  \bibinfo{year}{2022}.
\bibitem[{Liu et~al.(2023)Liu, Yang, Zhou, Liu, Wang, Liang, Tu, and Li}]{SCGC}
\bibinfo{author}{Y.~Liu}, \bibinfo{author}{X.~Yang}, \bibinfo{author}{S.~Zhou},
  \bibinfo{author}{X.~Liu}, \bibinfo{author}{S.~Wang},
  \bibinfo{author}{K.~Liang}, \bibinfo{author}{W.~Tu}, \bibinfo{author}{L.~Li},
\newblock \bibinfo{title}{Simple contrastive graph clustering},
\newblock \bibinfo{journal}{IEEE Transactions on Neural Networks and Learning
  Systems}  (\bibinfo{year}{2023}).
\bibitem[{Yang et~al.(2023)Yang, Liu, Zhou, Wang, Tu, Zheng, Liu, Fang, and
  Zhu}]{CCGC}
\bibinfo{author}{X.~Yang}, \bibinfo{author}{Y.~Liu}, \bibinfo{author}{S.~Zhou},
  \bibinfo{author}{S.~Wang}, \bibinfo{author}{W.~Tu},
  \bibinfo{author}{Q.~Zheng}, \bibinfo{author}{X.~Liu},
  \bibinfo{author}{L.~Fang}, \bibinfo{author}{E.~Zhu},
\newblock \bibinfo{title}{Cluster-guided contrastive graph clustering network},
\newblock in: \bibinfo{booktitle}{Proceedings of the AAAI conference on
  artificial intelligence}, volume~\bibinfo{volume}{37}, \bibinfo{year}{2023},
  pp. \bibinfo{pages}{10834--10842}.
\bibitem[{Liu et~al.(2022)Liu, Yang, Zhou, Liu, Wang, Liang, Tu, Li, Duan, and
  Chen}]{HSAN}
\bibinfo{author}{Y.~Liu}, \bibinfo{author}{X.~Yang}, \bibinfo{author}{S.~Zhou},
  \bibinfo{author}{X.~Liu}, \bibinfo{author}{Z.~Wang},
  \bibinfo{author}{K.~Liang}, \bibinfo{author}{W.~Tu}, \bibinfo{author}{L.~Li},
  \bibinfo{author}{J.~Duan}, \bibinfo{author}{C.~Chen},
\newblock \bibinfo{title}{Hard sample aware network for contrastive deep graph
  clustering},
\newblock \bibinfo{journal}{arXiv preprint arXiv:2212.08665}
  (\bibinfo{year}{2022}).
\bibitem[{Yang et~al.(2022)Yang, Liu, Zhou, Wang, Liu, and Zhu}]{GCC-LDA}
\bibinfo{author}{X.~Yang}, \bibinfo{author}{Y.~Liu}, \bibinfo{author}{S.~Zhou},
  \bibinfo{author}{S.~Wang}, \bibinfo{author}{X.~Liu},
  \bibinfo{author}{E.~Zhu},
\newblock \bibinfo{title}{Contrastive deep graph clustering with learnable
  augmentation},
\newblock \bibinfo{journal}{arXiv preprint arXiv:2212.03559}
  (\bibinfo{year}{2022}).
\bibitem[{Yang et~al.(2023)Yang, Tan, Liu, Liang, Wang, Zhou, Xia, Li, Liu, and
  Zhu}]{CONVERT}
\bibinfo{author}{X.~Yang}, \bibinfo{author}{C.~Tan}, \bibinfo{author}{Y.~Liu},
  \bibinfo{author}{K.~Liang}, \bibinfo{author}{S.~Wang},
  \bibinfo{author}{S.~Zhou}, \bibinfo{author}{J.~Xia}, \bibinfo{author}{S.~Z.
  Li}, \bibinfo{author}{X.~Liu}, \bibinfo{author}{E.~Zhu},
\newblock \bibinfo{title}{Convert: Contrastive graph clustering with reliable
  augmentation},
\newblock in: \bibinfo{booktitle}{Proceedings of the 31th ACM International
  Conference on Multimedia}, \bibinfo{year}{2023}.
\bibitem[{Hinton(2002)}]{nce1}
\bibinfo{author}{G.~E. Hinton},
\newblock \bibinfo{title}{Training products of experts by minimizing
  contrastive divergence},
\newblock \bibinfo{journal}{Neural computation} \bibinfo{volume}{14}
  (\bibinfo{year}{2002}) \bibinfo{pages}{1771--1800}.
\bibitem[{Hyv{\"a}rinen and Dayan(2005)}]{nce2}
\bibinfo{author}{A.~Hyv{\"a}rinen}, \bibinfo{author}{P.~Dayan},
\newblock \bibinfo{title}{Estimation of non-normalized statistical models by
  score matching.},
\newblock \bibinfo{journal}{Journal of Machine Learning Research}
  \bibinfo{volume}{6} (\bibinfo{year}{2005}).
\bibitem[{Van~den Oord et~al.(2018)Van~den Oord, Li, Vinyals et~al.}]{infonce}
\bibinfo{author}{A.~Van~den Oord}, \bibinfo{author}{Y.~Li},
  \bibinfo{author}{O.~Vinyals}, et~al.,
\newblock \bibinfo{title}{Representation learning with contrastive predictive
  coding},
\newblock \bibinfo{journal}{arXiv preprint arXiv:1807.03748}
  \bibinfo{volume}{2} (\bibinfo{year}{2018}) \bibinfo{pages}{4}.
\bibitem[{He et~al.(2020)He, Fan, Wu, Xie, and Girshick}]{MOCO}
\bibinfo{author}{K.~He}, \bibinfo{author}{H.~Fan}, \bibinfo{author}{Y.~Wu},
  \bibinfo{author}{S.~Xie}, \bibinfo{author}{R.~Girshick},
\newblock \bibinfo{title}{Momentum contrast for unsupervised visual
  representation learning},
\newblock in: \bibinfo{booktitle}{Proceedings of the IEEE/CVF Conference on
  Computer Vision and Pattern Recognition}, \bibinfo{year}{2020}, pp.
  \bibinfo{pages}{9729--9738}.
\bibitem[{Hassani and Khasahmadi(2020)}]{MVGRL}
\bibinfo{author}{K.~Hassani}, \bibinfo{author}{A.~H. Khasahmadi},
\newblock \bibinfo{title}{Contrastive multi-view representation learning on
  graphs},
\newblock in: \bibinfo{booktitle}{International Conference on Machine
  Learning}, \bibinfo{organization}{PMLR}, \bibinfo{year}{2020}, pp.
  \bibinfo{pages}{4116--4126}.
\bibitem[{Song et~al.(2018)Song, Zhang, Li, Gao, Wang, and Hong}]{autoencoder1}
\bibinfo{author}{J.~Song}, \bibinfo{author}{H.~Zhang}, \bibinfo{author}{X.~Li},
  \bibinfo{author}{L.~Gao}, \bibinfo{author}{M.~Wang},
  \bibinfo{author}{R.~Hong},
\newblock \bibinfo{title}{Self-supervised video hashing with hierarchical
  binary auto-encoder},
\newblock \bibinfo{journal}{IEEE Transactions on Image Processing}
  \bibinfo{volume}{27} (\bibinfo{year}{2018}) \bibinfo{pages}{3210--3221}.
\bibitem[{Hinton and Salakhutdinov(2006)}]{autoencoder2}
\bibinfo{author}{G.~E. Hinton}, \bibinfo{author}{R.~R. Salakhutdinov},
\newblock \bibinfo{title}{Reducing the dimensionality of data with neural
  networks},
\newblock \bibinfo{journal}{science} \bibinfo{volume}{313}
  (\bibinfo{year}{2006}) \bibinfo{pages}{504--507}.
\bibitem[{Kingma and Ba(2014)}]{ADAM}
\bibinfo{author}{D.~P. Kingma}, \bibinfo{author}{J.~Ba},
\newblock \bibinfo{title}{Adam: A method for stochastic optimization},
\newblock \bibinfo{journal}{arXiv preprint arXiv:1412.6980}
  (\bibinfo{year}{2014}).
\bibitem[{Wang et~al.(2021{\natexlab{a}})Wang, Liu, Zhu, Zhang, Zhang, Gao, and
  Zhu}]{siwei_1}
\bibinfo{author}{S.~Wang}, \bibinfo{author}{X.~Liu}, \bibinfo{author}{X.~Zhu},
  \bibinfo{author}{P.~Zhang}, \bibinfo{author}{Y.~Zhang},
  \bibinfo{author}{F.~Gao}, \bibinfo{author}{E.~Zhu},
\newblock \bibinfo{title}{Fast parameter-free multi-view subspace clustering
  with consensus anchor guidance},
\newblock \bibinfo{journal}{IEEE Transactions on Image Processing}
  \bibinfo{volume}{31} (\bibinfo{year}{2021}{\natexlab{a}})
  \bibinfo{pages}{556--568}.
\bibitem[{Wang et~al.(2021{\natexlab{b}})Wang, Liu, Liu, Zhou, and
  Zhu}]{siwei_2}
\bibinfo{author}{S.~Wang}, \bibinfo{author}{X.~Liu}, \bibinfo{author}{L.~Liu},
  \bibinfo{author}{S.~Zhou}, \bibinfo{author}{E.~Zhu},
\newblock \bibinfo{title}{Late fusion multiple kernel clustering with proxy
  graph refinement},
\newblock \bibinfo{journal}{IEEE Transactions on Neural Networks and Learning
  Systems}  (\bibinfo{year}{2021}{\natexlab{b}}).
\bibitem[{Zhang et~al.(2019)Zhang, Liu, and Fu}]{AE2Net}
\bibinfo{author}{C.~Zhang}, \bibinfo{author}{Y.~Liu}, \bibinfo{author}{H.~Fu},
\newblock \bibinfo{title}{Ae2-nets: Autoencoder in autoencoder networks},
\newblock in: \bibinfo{booktitle}{Proceedings of the IEEE/CVF conference on
  computer vision and pattern recognition}, \bibinfo{year}{2019}, pp.
  \bibinfo{pages}{2577--2585}.
\bibitem[{Xu et~al.(2021)Xu, Ren, Li, Pan, Zhu, and Xu}]{DEMVC}
\bibinfo{author}{J.~Xu}, \bibinfo{author}{Y.~Ren}, \bibinfo{author}{G.~Li},
  \bibinfo{author}{L.~Pan}, \bibinfo{author}{C.~Zhu}, \bibinfo{author}{Z.~Xu},
\newblock \bibinfo{title}{Deep embedded multi-view clustering with
  collaborative training},
\newblock \bibinfo{journal}{Information Sciences} \bibinfo{volume}{573}
  (\bibinfo{year}{2021}) \bibinfo{pages}{279--290}.
\bibitem[{Tang and Liu(2022)}]{DSMVC}
\bibinfo{author}{H.~Tang}, \bibinfo{author}{Y.~Liu},
\newblock \bibinfo{title}{Deep safe multi-view clustering: Reducing the risk of
  clustering performance degradation caused by view increase},
\newblock in: \bibinfo{booktitle}{Proceedings of the IEEE/CVF Conference on
  Computer Vision and Pattern Recognition}, \bibinfo{year}{2022}, pp.
  \bibinfo{pages}{202--211}.
\bibitem[{Van~der Maaten and Hinton(2008)}]{T_SNE}
\bibinfo{author}{L.~Van~der Maaten}, \bibinfo{author}{G.~Hinton},
\newblock \bibinfo{title}{Visualizing data using t-sne.},
\newblock \bibinfo{journal}{Journal of machine learning research}
  \bibinfo{volume}{9} (\bibinfo{year}{2008}).

\end{thebibliography}

\end{document}